%% file: neurips_2026.tex
\documentclass{article}

\PassOptionsToPackage{numbers,sort&compress}{natbib}

 \usepackage[main, final]{neurips_2026}
\usepackage[utf8]{inputenc} % allow utf-8 input
\usepackage[T1]{fontenc}    % use 8-bit T1 fonts
\usepackage{hyperref}       % hyperlinks
\usepackage{url}            % simple URL typesetting
\usepackage{booktabs}       % professional-quality tables
\usepackage{amsfonts}       % blackboard math symbols
\usepackage{nicefrac}       % compact symbols for 1/2, etc.
\usepackage{microtype}      % microtypography
\usepackage[table,dvipsnames]{xcolor}         % colors

\usepackage{amsmath}
\usepackage{enumerate} 
\usepackage{algorithm}
\usepackage{algpseudocode}
\usepackage{amsthm}
\usepackage{newtxtt}
\usepackage{diagbox} 
\usepackage{colortbl}
\usepackage{amssymb}
\usepackage{xspace}
\usepackage{wrapfig}
\usepackage{adjustbox}
\usepackage{placeins}
\usepackage{tabularx}
\usepackage{mathtools}
\usepackage{tikz}
\usepackage{enumitem}
\usepackage{silence}
\usepackage{dsfont}
\usepackage{multirow}
\usepackage{makecell}
\usepackage{xfakebold}
\usepackage{siunitx}
\usepackage{cleveref}
\usepackage{thmtools}

\input{math_commands}

\input{macros}

\usetikzlibrary{positioning, calc}
\usetikzlibrary{decorations.pathmorphing}

\newcommand{\MODEL}{LatentUM\xspace}

\title{LatentUM: Unleashing the Potential of \\ Interleaved Cross-Modal Reasoning via \\ a Latent-Space Unified Model}

\author{Jiachun Jin$^{1}$, Zetong Zhou$^{1}$, Xiao Yang$^{2}$, Hao Zhang$^{3}$, \\
  \textbf{Pengfei Liu$^{1}$, Jun Zhu$^{2}$, Zhijie Deng$^{1}$} \\
  $^1$Shanghai Jiao Tong University\quad
  $^2$Tsinghua University\quad
  $^3$UCSD \\
}

\begin{document}

\maketitle

\begin{abstract}
\input{sections/0_abstract}
\end{abstract}

\input{sections/1_intro}
\newpage
\input{sections/3_method}
\input{sections/4_exp}
\input{sections/2_related}

\input{sections/5_conclusion}
\begin{ack}
{This work was supported by the Shanghai Key Technology R\&D Program ``New Generation of Information Technology'' (No.~25511103700), and the National Natural Science Foundation of China (Nos.~62306176 and 92470118).}
\end{ack}

\FloatBarrier
\setlength{\bibsep}{5pt}
\bibliography{reference}
\bibliographystyle{plainnat}

%%%%%%%%%%%%%%%%%%%%%%%%%%%%%%%%%%%%%%%%%%%%%%%%%%%%%%%%%%%%

\appendix
\input{sections/apdx5_related}
\input{sections/apdx1_training_detail}
\input{sections/apdx2_self_reflection}
\input{sections/apdx3_fine_vsp_data}

\input{sections/apdx4_decoding_ref}
\input{sections/apdx6_zero-shot_language}
\input{sections/apdx7_additional_exp}

%%%%%%%%%%%%%%%%%%%%%%%%%%%%%%%%%%%%%%%%%%%%%%%%%%%%%%%%%%%%

\newpage
\input{checklist.tex}

\end{document}

%% file: math_commands.tex
\usepackage{amsmath,amsfonts,bm}

\def\eqref#1{equation~\ref{#1}}

\def\1{\bm{1}}

\DeclareMathAlphabet{\mathsfit}{\encodingdefault}{\sfdefault}{m}{sl}
\SetMathAlphabet{\mathsfit}{bold}{\encodingdefault}{\sfdefault}{bx}{n}

%% file: macros.tex
\usepackage{natbib}
\usepackage{amsfonts,bm}

\usepackage{amsmath, amssymb,mathrsfs, amsthm, mathtools}
\usepackage{xspace}
\usepackage{enumitem}
\usepackage{lipsum}
\usepackage{xpatch}
\usepackage{mathtools}
\usepackage{dsfont}
\usepackage{pgf,tikz}
\usetikzlibrary{positioning,matrix,patterns}
\usepackage{caption}
\usepackage{graphicx}
\usepackage{makecell}
\usepackage{multirow}
\usepackage[mathscr]{euscript}
\definecolor{hanblue}{rgb}{0.27, 0.42, 0.81}
\definecolor{deepred}{HTML}{900C3F}
\definecolor{deepgreen}{HTML}{2F6960}
\hypersetup{
    colorlinks=true,
    linkcolor=hanblue,
    urlcolor=hanblue,
    citecolor=hanblue,
    anchorcolor=hanblue}

\declaretheoremstyle[
  headfont=\sffamily\bfseries,
]{sansserif}

\theoremstyle{sansserif}

\theoremstyle{definition}

\theoremstyle{sansserif}

\theoremstyle{remark}

\DeclarePairedDelimiter\abs{\lvert}{\rvert}%
\DeclarePairedDelimiter\norm{\lVert}{\rVert}%

\makeatletter
\let\oldabs\abs
\def\abs{\@ifstar{\oldabs}{\oldabs*}}
\let\oldnorm\norm
\def\norm{\@ifstar{\oldnorm}{\oldnorm*}}
\makeatother

%% file: sections/0_abstract.tex
Unified models (UMs) hold promise for their ability to understand and generate content across heterogeneous modalities. Compared to merely generating visual content, the use of UMs for interleaved cross-modal reasoning is more promising and valuable, e.g., for solving understanding problems that require dense visual thinking, improving visual generation through self-reflection, or modeling visual dynamics of the physical world guided by stepwise action interventions. However, existing UMs typically rely on pixel decoding as a bridge when their visual representations for understanding and generation are disjoint, which is both ineffective and computationally cumbersome. In this paper, we introduce \textbf{LatentUM}, a novel unified model that represents all modalities within a shared semantic latent space, eliminating pixel-space mediation for model-internal reasoning over generated visual states. By making generated visual tokens directly interpretable to the model itself, LatentUM enables flexible interleaved cross-modal reasoning and generation. Empirically, LatentUM achieves state-of-the-art performance on the Visual Spatial Planning benchmark, pushes the limits of visual generation through self-reflection and demonstrates world modeling capability by predicting future visual states in the shared semantic latent space. Code and models are available at \url{https://github.com/SJTU-DENG-Lab/LatentUM}.

%% file: sections/1_intro.tex
\section{Introduction}
\begin{figure}[t]
    \centering
    \includegraphics[width=1\linewidth]{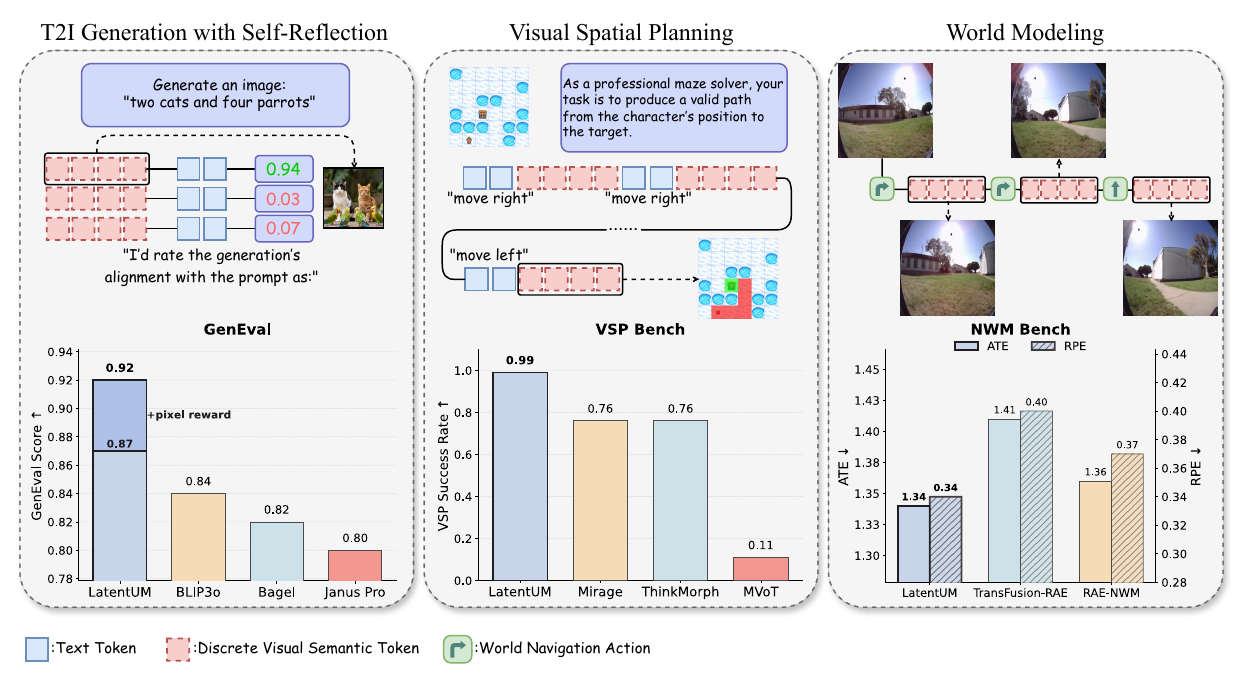}
    \caption{\textbf{\MODEL enables model-internal cross-modal reasoning in a shared semantic latent space.}
    \textbf{Left:} \MODEL improves text-to-image generation via self-reflection over its own generated semantic visual states.
    \textbf{Middle:} \MODEL interleaves textual reasoning with latent visual state predictions for visual spatial planning.
    \textbf{Right:} \MODEL predicts action-conditioned future visual latent states for world modeling.}
    \label{fig:teaser}
\end{figure}
Multimodal intelligence entails the seamless generation of diverse modalities, including text, images, and videos, with unified models (UMs) as a promising avenue for realizing this~\citep{team2024chameleon, chen2025janus, deng2025emerging, cui2025emu3}. 
However, existing UM approaches remain focused on visual generation tasks, such as image generation~\citep{chen2025blip3o, geng2025x, wu2025harmonizing, pan2025transfer}, image editing~\citep{deng2025emerging, wu2025omnigen2, li2025onecat, lin2025uniworld}, and video generation~\citep{xie2025show, wei2025univideo, ai2025ming}, leaving the broader potential of UMs for interleaved cross-modal reasoning largely underexplored.
Besides, the performance of UMs on these tasks also falls short of that of task-specialized models~\citep{wu2025qwen, wan2025, Qwen3-VL}.

This paper argues that the distinctive potential of UMs lies beyond standalone visual generation: they can serve as systems for interleaved cross-modal reasoning, e.g., solving visually grounded planning problems via step-by-step cross-modal reasoning,  improving visual generation through self-reflection, or performing world modeling based on stepwise actions.
These tasks are especially valuable for practical use, but place greater emphasis on semantic correctness than on pixel-level fidelity.
However, existing UMs are poorly suited to this setting when they maintain distinct visual representations for understanding and generation, requiring pixel decoding to serve as a bridge step~\citep{qin2025uni, gu2025thinkmorph}.
Such pixel-space mediation introduces unnecessary codec bias and cross-modal misalignment~\citep{yi2024bridge,fan2025prism}, leading to potentially degraded instruction-following ability on reasoning-centric tasks.

We propose \MODEL, which embeds various modalities within a semantic latent space, for cross-modal reasoning.
Motivated by the observation that reasoning-centric tasks emphasize semantic correctness over pixel-level fidelity, we advocate representing visual information as semantic tokens in the same space as language.
CLIP-style visual features~\citep{radford2021learning,tschannen2025siglip} provide a natural semantic representation, but their continuous nature (in contrast to discrete language tokens) can raise modeling complexity (e.g., requiring the involvement of diffusion modeling~\citep{kou2024orthus}).
We thus propose \textit{model behavior aligned quantization} (MBAQ) to discretize CLIP features into discrete visual semantic tokens, whose goal is to preserve the resultant vision-language prediction ability rather than pixel-level reconstruction fidelity. 
As a result, generated visual tokens become directly interpretable by \MODEL itself, removing the need for pixel-space mediation during model-internal reasoning.

Given the unified representation, \MODEL can be instantiated as an autoregressive (AR) transformer, employing a simple next-token prediction objective.
To preserve modality-specific capacity while enabling cross-modal interaction, we implement a \textit{Mixture-of-Modal Experts} (MoME) architecture, where modality-specific transformer block parameters process each modality and shared self-attention enables information exchange~\citep{vaswani2017attention,esser2024scaling,deng2025emerging}.
Benefiting from this design, \MODEL can directly inherit the visual understanding capability from off-the-shelf vision-language models (VLMs)~\citep{Qwen3-VL,wang2025internvl3} by utilizing them as parameter initialization. 
A diffusion decoder~\citep{peebles2023scalable,esser2024scaling} is additionally trained for pixel-space visualization when required, such as in text-to-image generation tasks.

We implement \MODEL{}\textsubscript{Base} from InternVL3.5-4B~\citep{wang2025internvl3} and train its visual generation branch on 32M text-to-image pairs~\citep{chen2025blip3o}.
Empirically, \MODEL{}\textsubscript{Base} preserves strong visual understanding while achieving competitive visual generation performance among existing unified models.
With task-specific post-training, \MODEL enables interleaved cross-modal reasoning for both visual generation with self-reflection and visually grounded planning, achieving state-of-the-art performance among unified models on GenEval~\citep{ghosh2023geneval}, GenEval2~\citep{kamath2025geneval}, and Visual Spatial Planning~\citep{wu2024vsp}.
We additionally extend \MODEL to action-conditioned world modeling~\citep{bar2025navigation,ha2018world}, where future visual states are predicted as discrete semantic tokens within the shared latent space, yielding semantically coherent rollouts and competitive performance against specialized world-modeling baselines.

%% file: sections/3_method.tex
\section{Method}
\label{sec:method}
Prevailing UMs cannot directly perform model-internal reasoning over generated visual states, because their generation and understanding pathways rely on separate visual representations~\citep{chen2025janus,deng2025emerging,xie2025show}. Generated states therefore have to be decoded into pixels and re-encoded for understanding, introducing codec bias and cross-modal misalignment~\citep{yi2024bridge}.
\MODEL eliminates this pixel-space mediation by representing both perceived and generated visual content within a shared semantic space.
Figure~\ref{fig:archi} presents an overview of \MODEL, which comprises three components: (i) a visual tokenizer implemented via model behavior aligned quantization (MBAQ); (ii) Mixture-of-Modal Experts (MoME) with modality-specific branches and shared self-attention; and (iii) a decoupled pixel decoder for optional visualization.

\subsection{Architecture}
\label{sec:architecture}
\begin{figure}[t]
    \centering
    \includegraphics[width=1\linewidth]{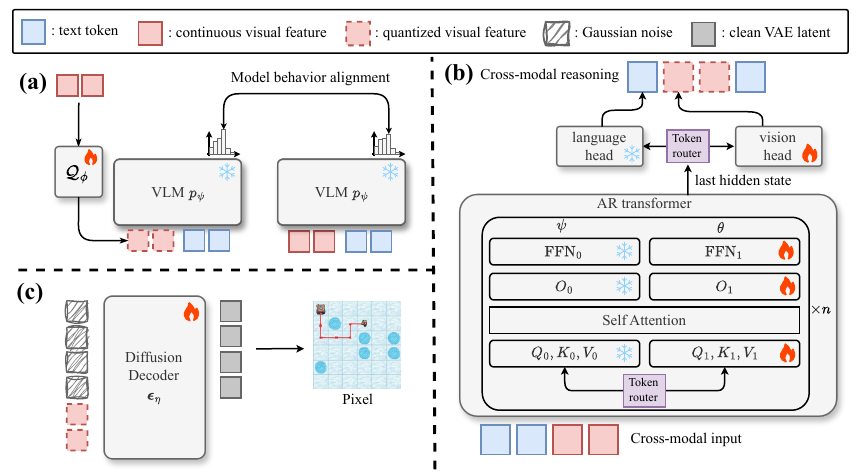}
    \caption{\textbf{Overview of \MODEL.}
    \MODEL unifies visual understanding and generation in a shared semantic latent space.
    \textbf{(a)} MBAQ converts CLIP features into discrete visual tokens by aligning with VLM behavior, preserving understanding-oriented semantics rather than pixel details.
    \textbf{(b)} Mixture-of-Modal Experts (MoME) decouples the understanding and generation branches while sharing self-attention for cross-modal reasoning.
    \textbf{(c)} An optional diffusion decoder renders semantic latents into pixels without training the core model for pixel reconstruction.}\label{fig:archi}
\end{figure}

\textbf{Visual Tokenizer.}
\label{sec:unified_repr}
Unlike prior work that relies on reconstruction-oriented visual features~\citep{kingma2013auto,esser2021taming} learned via pixel space recovery, \MODEL needs to generate understanding-oriented features for effective model-internal cross-modal reasoning.
A natural choice is to build upon CLIP features~\citep{radford2021learning}, which are inherently aligned with language through contrastive pretraining.
This alignment enables visual tokens to participate directly in cross-modal reasoning without pixel-space mediation.
Formally, let $\mathcal{I}$ denote an image and $\mathbf{V} \in \mathbb{R}^{d \times L}$ denote the corresponding CLIP features, where $L$ is the sequence length and $d$ is the channel dimension. 

However, the generative modeling of continuous CLIP features is non-trivial due to their high dimensionality (e.g., $d$ is typically on the order of thousands~\citep{zheng2025diffusion}). 
We can resort to extra diffusion modeling for such features on top of the autoregressive modeling of language, yet this can introduce substantial modeling and computational complexity. 
As a result, we decide to discretize $\mathbf{V}$ into discrete tokens for unified autoregressive modeling with language tokens.
Specifically, we leverage multi-codebook quantization (MCQ)~\citep{ma2025unitok}, which splits each $d$-dimensional token into $C$ chunks and quantizes them independently with separate codebooks of size $K$.
This yields an effective vocabulary of $K^C$ from only $K \times C$ codebook entries, producing a discrete code matrix $\mathbf{Z} \in [K]^{C \times L}$.

Regarding the quantization objective, we propose  \textit{model behavior aligned quantization} (MBAQ). 
The key insight is that quantized features should preserve the visual understanding capacities of the original ones, rather than merely reconstructing the latter. 
To this end, we incorporate a VLM $p_\psi$ that takes CLIP features as visual input, and enable it to also accept quantized visual features for visual understanding. 
Given a VQA dataset $\mathcal{D}_{\text{VQA}} = \{(\mathcal{I}^i, \mathbf{X}^i, \mathbf{Y}^i)\}_{i=1}^N$ with questions $\mathbf{X}^i$ and answers $\mathbf{Y}^i$ of length $\ell_i$, we train the quantizer $\mathcal{Q}_\phi$ by minimizing the KL divergence between the VLM's output distributions on $\mathbf{V}$ versus $\mathcal{Q}_\phi(\mathbf{V})$, as illustrated in Figure~\ref{fig:archi}(a). 
Specifically, the KL divergence at position $j$ of sample $i$ is:
\begin{equation}
    \mathcal{L}_{\text{KL}}^{i, j} = \mathbb{D}_{\textsf{KL}} \left[ p_{\psi}(\mathbf{Y}^i_j \mid \mathbf{Y}^i_{<j}, \mathbf{X}^i, \mathbf{V}^i) \parallel  p_{\psi}(\mathbf{Y}^i_j \mid \mathbf{Y}^i_{<j}, \mathbf{X}^i, \mathcal{Q}_{\phi}(\mathbf{V}^i)) \right].
\end{equation}
We denote the de-quantized features as $\tilde{\mathbf{V}} = \mathcal{Q}_{\phi}(\mathbf{V}) \in \mathbb{R}^{d \times L}$, which serve as a semantic approximation of the original CLIP features $\mathbf{V}$.
The full training objective averages this loss across all positions and samples, combined with the MCQ commitment loss $\mathcal{L}_{\text{MCQ}}$ weighted by $\lambda$.

\textbf{Mixture-of-Modal Experts (MoME).}
\label{sec:mome}
Given such visual tokens, we aim to build a unified autoregressive model to characterize the dependencies within and across modalities. 
However, naively performing unified training on a Transformer backbone can introduce conflicting optimization signals that degrade performance, as the language and visual generation tasks serve distinct purposes~\citep{song2025dualtoken, li2025unifork, hao2025uni}.

We address this issue with a Mixture-of-Modal Experts (MoME) architecture, as illustrated in Figure~\ref{fig:archi}(b).
The model maintains two parallel branches within each transformer layer: the understanding branch $\psi$ processes interleaved text and visual features for comprehension tasks, while the generation branch $\theta$ is dedicated to generating discrete visual tokens.
Such a decoupling draws an analogy with the Mixture-of-Transformer (MoT) architecture \citep{deng2025emerging}.

Within each layer, the two branches share the self-attention mechanism while maintaining separate feed-forward networks and projection matrices.
This shared attention enables cross-modal information flow, allowing the generation branch to leverage understanding context.
A simple routing mechanism enables seamless switching between understanding and generation modes at inference time.
Routing is determined by a special \texttt{<BOI>} token: tokens following \texttt{<BOI>} are processed by $\theta$ to produce discrete visual codes; otherwise, $\psi$ handles language generation.

For the prediction head, we implement a lightweight causal transformer that predicts the $C$ code indices sequentially within each visual token~\citep{wang2025bridging,ma2025unitok}.
This design captures the dependencies among code indices within the same token position, which a simple linear classification head would ignore~\citep{ma2025unitok}.
We provide further details on the pretraining procedure in Section~\ref{sec:pretraining}.

\textbf{Decoupled Pixel Decoder.}
\label{sec:decoder}
Although the visual features in \MODEL are optimized for semantic reasoning rather than pixel reconstruction, they can be rendered into images by a separate diffusion decoder $\boldsymbol{\epsilon}_\eta$.
We adapt a pretrained text-to-image diffusion transformer~\citep{peebles2023scalable,esser2024scaling} by replacing its text conditioning with the quantized visual features $\tilde{\mathbf{V}}$, and train it with a standard conditional denoising objective:
\begin{equation}\label{eq:diff}
    \mathcal{L}_{\text{diff}}(\eta) = \mathbb{E}_{t, \boldsymbol{\epsilon}, \mathbf{x}_0}
    \left[ \| \boldsymbol{\epsilon} - \boldsymbol{\epsilon}_\eta(\mathbf{x}_t, t, \tilde{\mathbf{V}}) \|^2 \right],
\end{equation}
where $\mathbf{x}_t$ is the noisy latent at timestep $t$, and $\boldsymbol{\epsilon}$ is the added noise. 

\begin{figure}[t]
    \centering
    \begin{minipage}[t]{0.66\textwidth}
        \centering
        \includegraphics[width=\linewidth]{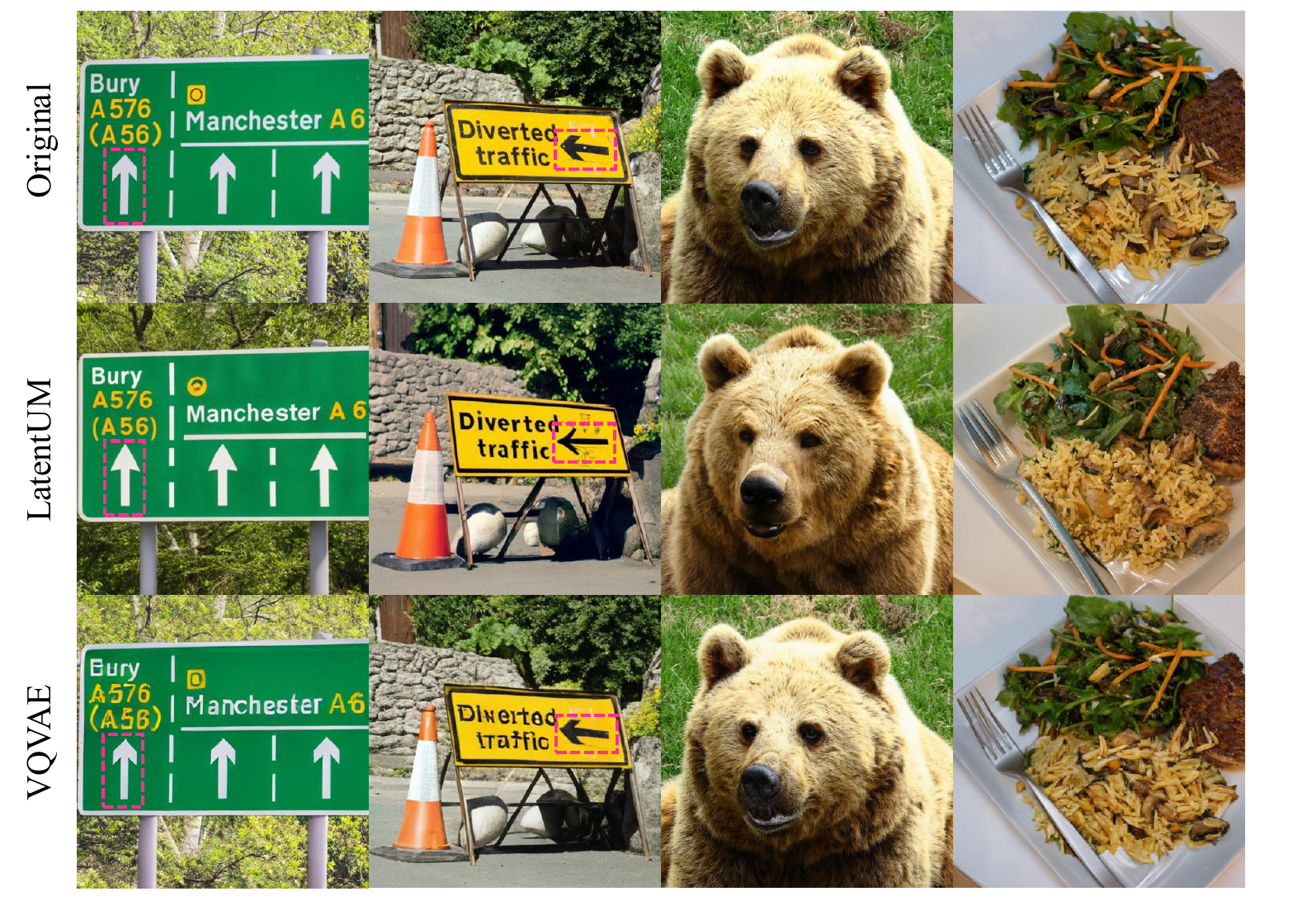}
    \end{minipage}
    \hfill
    \begin{minipage}[t]{0.32\textwidth}
        \centering
        \includegraphics[width=\linewidth]{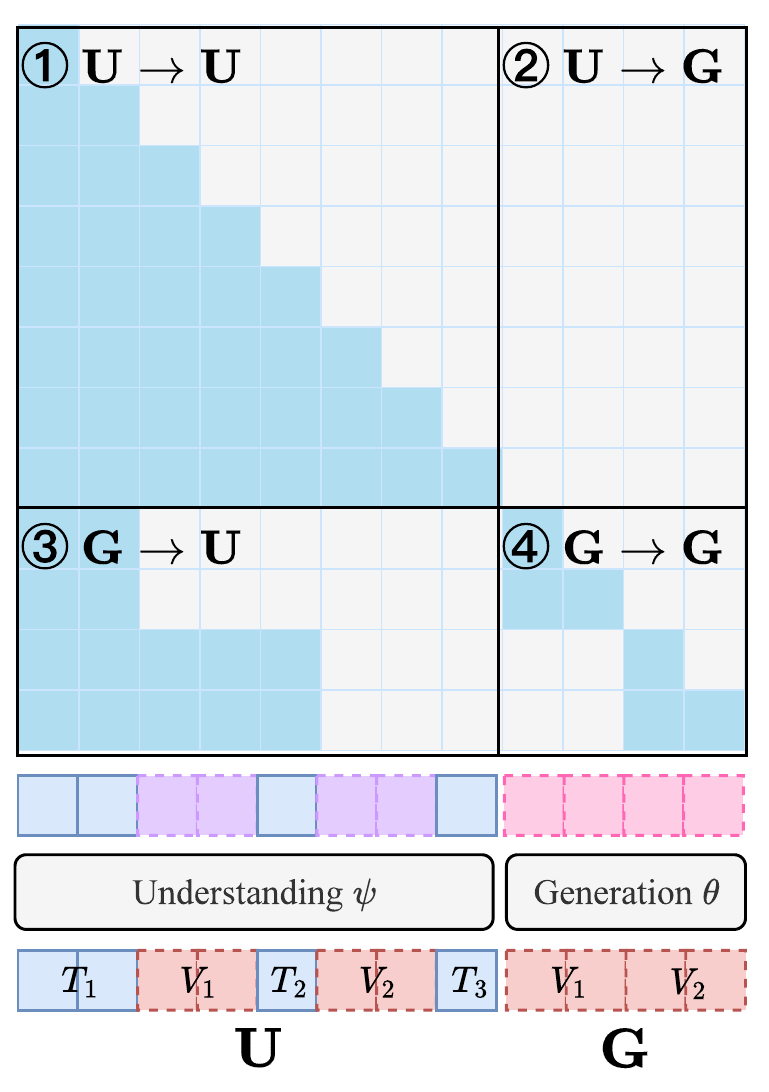}
    \end{minipage}
    \caption{
    \textbf{Left: Pixel reconstruction comparison.} VQVAE preserves low-level details such as arrow styles but loses semantic content such as sign text, whereas our quantizer preserves semantics while discarding non-essential pixel details.
    \textbf{Right: Multi-frame interleaved training.} A dedicated attention mask routes visual tokens through both MoME branches, allowing all visual states to be trained in one forward pass.}
    \label{fig:pretrain_aux}
\end{figure}

The decoder is trained independently, so the core model never optimizes for pixel reconstruction. As shown in Figure~\ref{fig:pretrain_aux} (left), MBAQ features retain enough semantics for faithful rendering while discarding fine-grained pixel details less relevant to understanding, unlike VQVAE-style features.

\subsection{Pre-training}
\label{sec:pretraining}

For MBAQ training, we use the CLIP encoder of InternVL3.5-4B~\citep{wang2025internvl3,tschannen2025siglip} and train $\mathcal{Q}_\phi$ on LLaVA-v1.5-665K~\citep{liu2024improved}. We instantiate \MODEL{}\textsubscript{Base} from InternVL3.5-4B, initialize the understanding branch $\psi$ from its pretrained weights, thereby directly inheriting its strong visual understanding ability, and train only the generation branch $\theta$ on 32M BLIP3o image-text pairs~\citep{chen2025blip3o} with a multi-code next-token prediction objective:
\begin{equation}
    \label{eq:mcntp}
    \mathcal{L}_{\text{MC-NTP}}(\theta) =
    - \frac{1}{L} \sum_{\ell=1}^{L} \frac{1}{C} \sum_{c=1}^{C}
    \log p_{\theta}(\mathbf{Z}_{c, \ell} \mid \mathbf{Z}_{<c, <\ell}, \mathbf{X}),
\end{equation}
where $\mathbf{Z}_{c, \ell}$ denotes the $c$-th code index of the $\ell$-th visual token, and $\mathbf{X}$ denotes the conditioning context. Pre-training uses text-only contexts due to data availability, while interleaved cross-modal conditioning is activated during post-training.

We train the diffusion decoder $\boldsymbol{\epsilon}_\eta$ independently on the same 32M images by fine-tuning the condition branch of a pretrained dual-stream diffusion transformer~\citep{esser2024scaling}. Detailed training settings are provided in Appendix~\ref{app:pretraining}.

\subsection{Post-training}
\label{sec:posttraining}
Although \MODEL{}\textsubscript{Base} possesses the architectural capacity for model-internal cross-modal reasoning, activating it requires task-specific post-training.

\textbf{Supervised Fine-tuning.}
To enable \emph{cross-modal chain-of-thought} reasoning, we jointly fine-tune the understanding branch $\psi$ and generation branch $\theta$ on interleaved cross-modal sequences with a next-token prediction objective:
\begin{equation}
    \label{eq:sft}
    \mathcal{L}_{\text{SFT}}(\psi, \theta) =
    - \sum_{i=1}^{L} \log p_{\Theta}(\mathbf{o}_i \mid \mathbf{o}_{<i}, \mathbf{X}),
\end{equation}
where $\Theta \in \{\psi, \theta\}$ selects the branch according to token type, $\mathbf{o}_i$ denotes the $i$-th token, and $\mathbf{X}$ is the interleaved multimodal context. Generated visual tokens are re-processed by $\psi$ and cached as understanding context for subsequent tokens. During SFT, each visual segment is therefore routed through both $\theta$ and $\psi$ with a parallel attention mask, enabling generation loss and understanding-context construction in one forward pass as illustrated in Figure~\ref{fig:pretrain_aux} (right).

\textbf{Reinforcement Learning.}
Given a prompt, \MODEL{} samples $G$ cross-modal rollouts $\{\mathbf{O}_g\}_{g=1}^G$, where each response may contain both text and visual tokens. We can optimize \MODEL{} with the following GRPO~\citep{shao2024deepseekmath} objective:
\begin{equation}
    \mathcal{L}_{\text{GRPO}}(\psi, \theta) = \mathbb{E}_{\{\mathbf{O}_g\}} \left[ \frac{1}{G} \sum_{g=1}^G \frac{1}{L_g} \sum_{i=1}^{L_g}
    \min \bigl( r_g^i \hat{A}_g, \, \text{clip}(r_g^i) \hat{A}_g \bigr)
    - \beta \, \mathbb{D}_{\textsf{KL}} [ p_{\Theta} \| p_{\text{ref}} ] \right].
\end{equation}
Here $\Theta \in \{\psi, \theta\}$ selects the branch according to token type, $r_g^i = p_{\Theta}(\mathbf{o}_i) / p_{\text{old}}(\mathbf{o}_i)$ is the clipped probability ratio, and $\hat{A}_g = (\mathcal{R}_g - \bar{\mathcal{R}}) / \sigma_{\mathcal{R}}$ is the normalized group advantage.
The reward $\mathcal{R}_g$ can be provided by external reward models or by the understanding branch $\psi$ itself via \emph{generate-then-reflect}. In the latter case, $\psi$ acts as a semantic evaluator over generated visual features and returns a scalar reward. Our experiments instantiate this reward with verification questions, as described in Section~\ref{sec:exp_reason_gen} and Appendix~\ref{app:self_reflection_analysis}.

%% file: sections/4_exp.tex
\section{Experiments}
\label{sec:experiment}
In this section, we first assess the basic visual understanding and generation capabilities of \MODEL{}\textsubscript{Base} in Section~\ref{sec:exp_base_model}, then study \MODEL's model-internal cross-modal reasoning abilities in Section~\ref{sec:cross_modal_reasoning}, extend the evaluation to action-conditioned world modeling in Section~\ref{sec:world_modeling}, and finally ablate the semantic latent-space design choices in Section~\ref{sec:ablation}. Detailed experimental setup is provided in Appendix~\ref{app:exp_setup}.
% We comprehensively evaluate \MODEL on visual understanding, visual generation, and cross-modal reasoning tasks. Our experiments aim to answer four questions: (1) How does \MODEL perform on standard visual understanding and generation benchmarks compared to existing unified models? (2) Can \MODEL leverage its own understanding capability to enhance visual generation? (3) Can \MODEL reason over its own generated visual representations to solve complex planning tasks? (4) Can \MODEL perform world modeling?

\subsection{Base Model Capabilities}
\label{sec:exp_base_model}
\begin{figure*}[h]
    \centering
    \includegraphics[width=\textwidth]{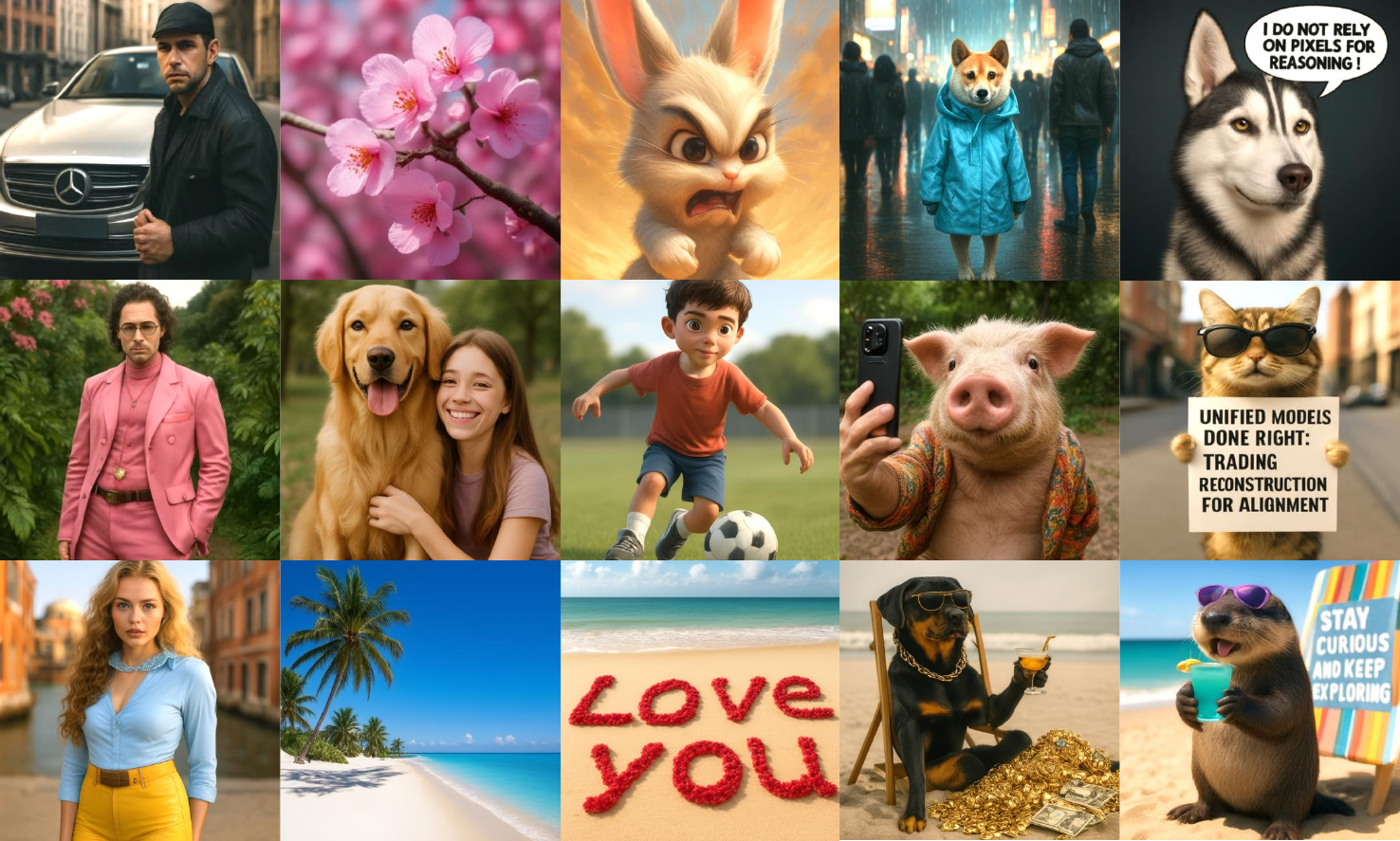}
    \caption{\textbf{Text-to-image gallery of \MODEL.} The last column shows legible in-image text enabled by the unified semantic space.}
    \label{fig:demo}
\end{figure*}
\begin{table}[!h]
    \centering
    % \captionsetup{font=small,skip=2pt}
    \begin{minipage}[t]{0.45\textwidth}
        \centering
        \caption{Evaluation results on multimodal understanding benchmarks. $^\dagger$ denotes evaluation with quantized visual features.}
        \label{tab:visual_under}
        \tiny
        \setlength{\tabcolsep}{1.8pt}
        \begin{adjustbox}{max width=\linewidth}
            \begin{tabular}{@{}lccccc@{}}
                \toprule
                Model & MME-P $\uparrow$ & POPE $\uparrow$ & SEED $\uparrow$ & MMB $\uparrow$ & MMMU $\uparrow$ \\
                \midrule
                Tokenflow XL~\citep{qu2025tokenflow} & 1551 & 87.8 & 72.6 & 76.8 & 43.2 \\
                Bagel~\citep{deng2025emerging} & 1687 & - & - & 85.0 & 55.3 \\
                JanusFlow~\citep{ma2025janusflow} & 1333 & 88.0 & 70.5 & - & - \\
                Harmon~\citep{wu2025harmonizing} & 1155 & 87.6 & 67.1 & 65.5 & - \\
                Chameleon~\citep{team2024chameleon} & 1057 & 77.8 & - & - & 26.7 \\
                Orthus~\citep{kou2024orthus} & 1266 & 79.6 & - & - & 28.2 \\
                EMU3~\citep{wang2024emu3} & 1244 & 85.2 & 68.2 & 58.5 & - \\
                VILA-U~\citep{wu2024vila} & 1402 & 85.8 & 59.0 & - & - \\
                Janus Pro~\citep{chen2025janus} & 1444 & 87.4 & 72.1 & 79.2 & 41.0 \\
                Show-o2~\citep{xie2025show} & 1620 & - & 69.8 & 79.3 & 48.9 \\
                BLIP3o 4B~\citep{chen2025blip3o} & 1528 & - & 73.8 & 78.6 & 46.6 \\
                BLIP3o 8B~\citep{chen2025blip3o} & 1683 & - & 77.5 & 83.5 & 50.6 \\
                InternVL3.5 4B~\citep{wang2025internvl3} & 1654 & 88.9 & 76.3 & 80.3 & 54.6 \\
                \midrule
                \rowcolor[HTML]{D9F3FD}
                \MODEL{}\textsubscript{Base} & 1654 & 88.9 & 76.3 & 80.3 & 54.6 \\
                \rowcolor[HTML]{D9F3FD}
                \MODEL{}\textsubscript{Base}$^\dagger$ & 1638 & 85.5 & 75.0 & 79.5 & 52.3 \\
                \bottomrule
            \end{tabular}
        \end{adjustbox}
    \end{minipage}
    \hfill
    \begin{minipage}[t]{0.53\textwidth}
        \centering
        \caption{Evaluation results on GenEval~\citep{ghosh2023geneval}. \textit{+ pixel-reward} denotes using an external pixel-reward model.}
        \label{tab:geneval}
        \tiny
        \setlength{\tabcolsep}{3pt}
        \begin{adjustbox}{max width=\linewidth}
            \begin{tabular}{@{}lccccccc@{}}
                \toprule
                Model & Single & Two & Count & Color & Pos. & Color Attr. & Overall $\uparrow$ \\
                \midrule
                \multicolumn{8}{l}{\textit{Specialized Text-to-Image Models}} \\
                FLUX.1 [Dev]~\citep{flux2024} & 0.98 & 0.81 & 0.74 & 0.79 & 0.22 & 0.45 & 0.66 \\
                SD3.5 Large~\citep{esser2024scaling} & 0.98 & 0.89 & 0.73 & 0.83 & 0.34 & 0.47 & 0.71 \\
                Qwen-Image~\citep{wu2025qwen} & 0.99 & 0.92 & 0.89 & 0.88 & 0.76 & 0.77 & 0.87 \\
                \midrule
                \multicolumn{8}{l}{\textit{Unified Models}} \\
                TokenFlow-XL~\citep{qu2025tokenflow} & 0.95 & 0.60 & 0.41 & 0.81 & 0.16 & 0.24 & 0.55 \\
                Transfusion~\citep{zhou2024transfusion} & - & - & - & - & - & - & 0.63 \\
                JanusFlow~\citep{ma2025janusflow} & 0.97 & 0.59 & 0.45 & 0.83 & 0.53 & 0.42 & 0.63 \\
                Bagel~\citep{deng2025emerging} & 0.99 & 0.94 & 0.81 & 0.88 & 0.64 & 0.63 & 0.82 \\
                Janus Pro~\citep{chen2025janus} & 0.99 & 0.89 & 0.59 & 0.90 & 0.79 & 0.66 & 0.80 \\
                Show-o2~\citep{xie2025show} & 1.00 & 0.87 & 0.58 & 0.92 & 0.52 & 0.62 & 0.76 \\
                BLIP3o 4B~\citep{chen2025blip3o} & - & - & - & - & - & - & 0.81 \\
                BLIP3o 8B~\citep{chen2025blip3o} & - & - & - & - & - & - & 0.84 \\
                \midrule
                \rowcolor[HTML]{D9F3FD}
                \MODEL{}\textsubscript{Base} & 0.99 & 0.92 & 0.72 & 0.91 & 0.83 & 0.73 & 0.85 \\
                \rowcolor[HTML]{D9F3FD}
                \MODEL{}\textsubscript{Vis-Gen} & 0.99 & 0.92 & 0.85 & 0.87 & 0.89 & 0.71 & 0.87 \\
                \rowcolor[HTML]{D9F3FD}
                + \textit{pixel-reward} & 0.99 & 0.95 & 0.95 & 0.92 & 0.92 & 0.81 & 0.92 \\
                \bottomrule
            \end{tabular}
        \end{adjustbox}
    \end{minipage}
\end{table}
\textbf{Visual Understanding.}
We evaluate \MODEL{}\textsubscript{Base}'s visual understanding on MME~\citep{fu2025mmecomprehensiveevaluationbenchmark}, POPE~\citep{li2023evaluating}, SEED-Bench~\citep{li2023seed}, MM-Bench~\citep{liu2024mmbench}, and MMMU~\citep{yue2024mmmu}, using both continuous and quantized visual features.
As shown in Table~\ref{tab:visual_under}, \MODEL{}\textsubscript{Base} retains strong visual understanding by directly inheriting the frozen InternVL3.5 understanding branch. Quantization causes only modest degradation, indicating that MBAQ produces semantically meaningful representations, which is a prerequisite for effective cross-modal reasoning over self-generated visual content.

\textbf{Visual Generation.}
Figure~\ref{fig:demo} shows semantically coherent text-to-image samples from \MODEL{}\textsubscript{Base}, including emergent in-image text rendering.
On GenEval~\citep{ghosh2023geneval}, \MODEL{}\textsubscript{Base} achieves an overall score of 0.85, outperforming all unified baselines in Table~\ref{tab:geneval}. These results suggest that MBAQ-aligned visual features support strong instruction following in visual generation.

\subsection{Cross-Modal Reasoning}
\label{sec:cross_modal_reasoning}

\textbf{Visual Generation with Self-Reflection.}
\label{sec:exp_reason_gen}
\MODEL can improve its visual generation quality through \emph{generate-then-reflect}, a model-internal reasoning process where the model reflects on its own generated semantic visual states without decoding them back to pixels. Following the RL framework in Section~\ref{sec:posttraining}, we post-train \MODEL{}\textsubscript{Base} with self-reward derived from its own understanding capability, yielding \MODEL{}\textsubscript{Vis-Gen}.

For GenEval~\citep{ghosh2023geneval} and GenEval2~\citep{kamath2025geneval}, we decompose each training prompt (disjoint from the evaluation prompts) into atomic visual checks, such as object existence, counts, colors, and spatial relations, and use the resulting verification questions to compute self-reward with the understanding branch directly over generated semantic visual features. This avoids decoding the generated visual state into pixels or invoking an external pixel-space reward model. We keep $\psi$ fixed and fine-tune only the visual-generation pathway, decoupling reward evaluation from the learned policy. Detailed question templates, option scoring, reward aggregation, and GRPO settings are provided in Appendix~\ref{app:self_reflection_analysis}.

As shown in Table~\ref{tab:geneval}, \MODEL{}\textsubscript{Vis-Gen} achieves 0.87 on GenEval, outperforming all unified model baselines; an external  GenEval pixel-reward model~\citep{liu2025flow, zheng2025diffusionnft} further raises the score to 0.92.
This gap more likely reflects limitations of GenEval's detection-based metric than a weakness of model-internal self-reward, as valid generations can be penalized by detector false negatives; see Appendix~\ref{sec:failure_detection_metric}.
In contrast, GenEval2~\citep{kamath2025geneval} employs MLLM-as-judge evaluation~\citep{chen2024mllm}, which is more semantically robust; there, as shown in Table~\ref{tab:geneval2}, \MODEL{}\textsubscript{Vis-Gen} outperforms all baselines by a large margin, confirming the effectiveness of model-internal self-reward.
\begin{figure*}[t]
    \centering
    \includegraphics[width=\linewidth]{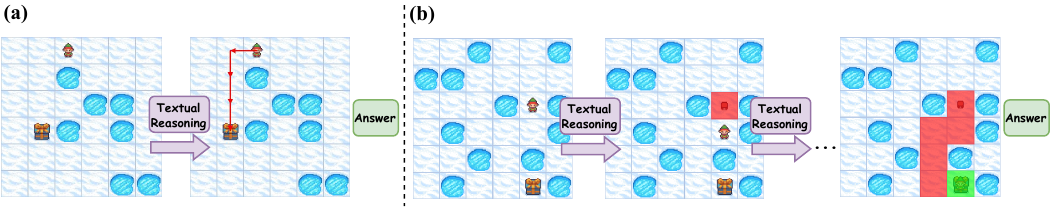}
    \caption{\textbf{Two interleaved reasoning paradigms for visual spatial
    planning.}
    \textbf{(a)} Coarse-grained: generates a visual plan from the maze layout
    before solving.
    \textbf{(b)} Fine-grained: alternates textual reasoning with visual state predictions to guide each action.}
    \label{fig:vsp}
\end{figure*}
\begin{table}[t]
    \centering
    \begin{minipage}[t]{0.47\textwidth}
        \centering
        \caption{Evaluation results on GenEval2~\citep{kamath2025geneval}.}
        \label{tab:geneval2}
        \footnotesize
        \begin{tabular}{@{}lcc@{}}
            \toprule
            Model & TIFA\textsubscript{GM} & TIFA\textsubscript{AM} \\
            \midrule
            Bagel + CoT~\citep{deng2025emerging} & 23.1 & 70.9 \\
            Janus Pro~\citep{chen2025janus} & 14.5 & 58.3 \\
            BLIP3o 8B~\citep{chen2025blip3o} & 13.3 & 59.4 \\
            \midrule
            \rowcolor[HTML]{D9F3FD}
            \MODEL{}\textsubscript{Base} & 21.4 & 68.1 \\
            \rowcolor[HTML]{D9F3FD}
            \MODEL{}\textsubscript{Vis-Gen} & 31.3 & 72.9 \\
            \bottomrule
        \end{tabular}
    \end{minipage}
    \hfill
    \begin{minipage}[t]{0.5\textwidth}
        \centering
        \caption{Evaluation results on VSP~\citep{wu2024vsp}.}
        \label{tab:vsp}
        \footnotesize
        \setlength{\tabcolsep}{4pt}
        \begin{tabular}{@{}lccccc@{}}
            \toprule
            Model & L3 & L4 & L5 & L6 & Avg. \\
            \midrule
            Anole~\citep{chern2024anole} & 0.02 & 0.01 & 0.00 & 0.00 & 0.01 \\
            MVoT~\citep{li2025imagine} & 0.21 & 0.11 & 0.08 & 0.03 & 0.11 \\
            Mirage~\citep{yang2025machine} & 0.93 & 0.83 & 0.76 & 0.51 & 0.76 \\
            Chameleon~\citep{team2024chameleon} & - & - & - & - & 0.01 \\
            InternVL3.5 8B~\citep{wang2025internvl3} & - & - & - & - & 0.08 \\
            InternVL3.5 38B~\citep{wang2025internvl3} & - & - & - & - & 0.20 \\
            ThinkMorph~\citep{gu2025thinkmorph} & - & - & - & - & 0.76 \\
            \midrule
            \rowcolor[HTML]{D9F3FD}
            \MODEL{}\textsubscript{Vis-Plan} (coarse) & 1.00 & 0.85 & 0.83 & 0.71 & 0.85 \\
            \rowcolor[HTML]{D9F3FD}
            \MODEL{}\textsubscript{Vis-Plan} (fine) & 1.00 & 1.00 & 1.00 & 0.97 & 0.99 \\
            \bottomrule
        \end{tabular}
    \end{minipage}
\end{table}

\textbf{Interleaved Visual Reasoning for Spatial Planning.}
\label{sec:exp_reason_under}
A key advantage of \MODEL's unified latent space is enabling model-internal reasoning over generated visual states to solve visual reasoning tasks.
We evaluate this capability on the Visual Spatial Planning (VSP) benchmark~\citep{wu2024vsp}, which requires spatial reasoning in a maze-navigation environment.
We fine-tune \MODEL{}\textsubscript{Base} on task-specific data for each paradigm, yielding \MODEL{}\textsubscript{Vis-Plan}.
We investigate two reasoning paradigms that exploit this capability in progressively deeper ways, as illustrated in Figure~\ref{fig:vsp}.

\noindent\textit{Coarse-grained planning.}
This follows a text-visual-text reasoning flow: it first analyzes the maze layout in text, then generates a complete visual plan, and finally derives the textual solution from this plan. The generated plan serves as an intermediate visual reasoning state in the shared latent space. We use the 6{,}000 training samples from ThinkMorph~\citep{gu2025thinkmorph} in the format of Figure~\ref{fig:vsp}(a).

\noindent\textit{Fine-grained step-by-step planning.}
This repeats a text-visual-text loop: each textual reasoning produces an updated visual state. This stepwise interleaving yields denser cross-modal context throughout the reasoning trajectory. We construct 18{,}000 filtered training samples in the format of Figure~\ref{fig:vsp}(b), with details provided in Appendix~\ref{app:vsp_data}.

As reported in Table~\ref{tab:vsp}, \MODEL{}\textsubscript{Vis-Plan} outperforms prior visual reasoning models under both planning paradigms, including the directly comparable ThinkMorph baseline~\citep{gu2025thinkmorph}, which is fine-tuned from Bagel~\citep{deng2025emerging} with the same data and relies on pixel-space mediation.
These gains validate our core hypothesis: unified semantic representations enable more effective model-internal cross-modal reasoning than pixel-space bridging, while the coarse-to-fine improvement shows that the shared latent space supports deeper interleaved reasoning for complex spatial planning.
% As reported in Table~\ref{tab:vsp}, \MODEL{}\textsubscript{Vis-Plan} consistently outperforms all prior visual reasoning models across both planning paradigms.
% Of particular note is the comparison with ThinkMorph~\citep{gu2025thinkmorph}, which is fine-tuned from Bagel~\citep{deng2025emerging} and relies on pixel-space mediation for cross-modal reasoning, making it a directly comparable baseline with \MODEL{}\textsubscript{Vis-Plan}.
% The substantial performance gap in favor of \MODEL validates our core hypothesis: unified semantic representations enable more effective cross-modal reasoning than approaches that require pixel-space bridging. Furthermore, the transition from coarse-grained to fine-grained planning yields significant improvements, demonstrating that the unified latent representation supports deeper interleaved reasoning patterns for complex spatial planning tasks.

\subsection{World Modeling}
\label{sec:world_modeling}
\begin{figure}[t]
    \centering
    \includegraphics[width=\linewidth]{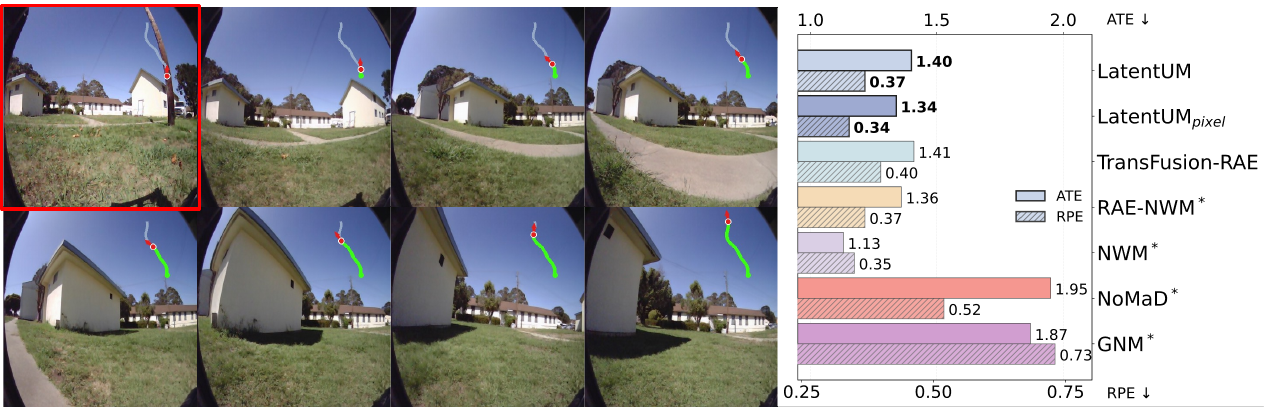}
    \caption{
    \textbf{Left:} Open-loop future-frame rollouts following a predefined action trajectory.
    The red box denotes the initial context frame, and subsequent frames are predicted future observations.
    \textbf{Right:} Quantitative comparison on the NWM benchmark~\citep{bar2025navigation}. $^*$ denotes specialist for navigation.}
    \label{fig:world_modeling}
\end{figure}
Beyond reasoning over static generated visual states, we ask whether the same latent space can support temporally grounded reasoning: predicting future visual states conditioned on actions~\citep{ha2018world}.

We adopt the Navigation World Model (NWM)~\citep{bar2025navigation} evaluation setting and fine-tune \MODEL{}\textsubscript{Base} on egocentric robot navigation dataset RECON~\citep{shah2021rapid}, to obtain \MODEL{}\textsubscript{WM}. Following Transfusion-RAE~\citep{tong2026beyond}, we represent navigation actions as standard text tokens, seamlessly integrating into \MODEL's unified autoregressive formulation. Each training sample is constructed by conditioning on four context frames and a text-based navigation action to predict the discrete semantic tokens of the subsequent frame. During inference, \MODEL supports fully latent rollout by iteratively feeding predicted semantic codes as subsequent context. Since the NWM evaluation is sensitive to fine-grained pixel-space structure, we also evaluate a pixel-space recurrent interface that renders predicted semantic codes into images and re-encodes them at each step, while keeping latent semantic-token prediction as the learned dynamics model.

To validate the learned dynamics, we perform open-loop future-frame prediction: Figure~\ref{fig:world_modeling} (left) shows temporally consistent rollouts that follow a predefined action trajectory while preserving the initial scene layout.
Notably, due to the inherited language-vision alignment, \MODEL{}\textsubscript{WM} also exhibits an emergent capability of zero-shot language-conditioned navigation, enabling it to natively follow free-form textual instructions, as detailed in Appendix~\ref{app:wm_lang}.
Furthermore, we evaluate under the NWM zero-shot planning protocol.
As shown in Figure~\ref{fig:world_modeling} (right), fully latent rollout achieves 1.40 ATE / 0.37 RPE, while pixel-space recurrence improves this to 1.34 / 0.34, surpassing Transfusion-RAE~\citep{tong2026beyond} and remaining competitive with the specialized NWM baseline. These results suggest that the shared semantic latent space extends beyond static reasoning to action-conditioned future-state prediction, offering a preliminary step toward temporally grounded world modeling.

\subsection{Ablation Studies}
\label{sec:ablation}

We conduct ablation studies to validate two core design choices in \MODEL: (i) using semantically aligned features instead of pixel-reconstruction features for visual generation, and (ii) training the quantizer with MBAQ rather than feature reconstruction objectives for visual understanding.

\begin{figure}[t]
    \centering
    \includegraphics[width=\linewidth]{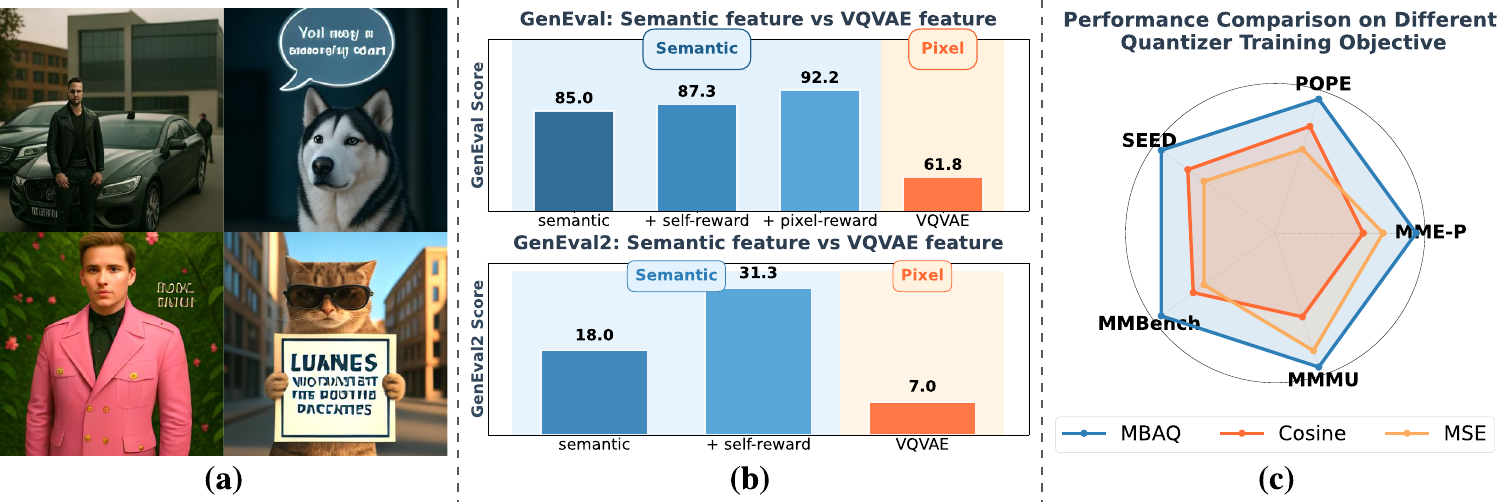}
    \caption{\textbf{Ablation study.} \textbf{(a)} Text-to-image generation with the VQVAE baseline on selected prompts from Figure~\ref{fig:demo}.
    \textbf{(b)} GenEval and GenEval2 comparison, showing stronger instruction following from semantically aligned features.
    \textbf{(c)} Visual understanding comparison across quantizer training objectives, where MBAQ consistently outperforms feature-reconstruction baselines.}
    \label{fig:ablation}
\end{figure}

\textbf{Semantic Features for Visual Generation.}
\label{sec:aba_llamagen}
We first validate that semantically aligned features outperform pixel-reconstruction features for instruction-following in visual generation.
We train an alternative model that differs from \MODEL \emph{only} in the visual features: it adopts VQVAE features from LlamaGen~\citep{sun2024autoregressive}, while architecture and training data remain identical.
Figure~\ref{fig:ablation}(a) shows qualitative results. The VQVAE baseline generates plausible images but exhibits a clear quality gap compared to \MODEL, particularly in text rendering. Figure~\ref{fig:ablation}(b) quantifies this gap: \MODEL outperforms the baseline on both GenEval and GenEval2, confirming that semantically aligned visual features are crucial for instruction-following in visual generation.

\textbf{MBAQ for Visual Understanding.}
\label{sec:aba_mbaq}
We next investigate how different quantizer training objectives affect visual understanding.
We compare MBAQ against two alternative quantizers that directly reconstruct continuous semantic features using either MSE loss or cosine similarity loss~\citep{han2506vision}, sharing the same architecture. As shown in Figure~\ref{fig:ablation}(c), MBAQ consistently outperforms both reconstruction-based alternatives across five visual understanding benchmarks. This indicates that incorporating next-token prediction behavior during quantizer training is more effective than approaches that focus solely on feature reconstruction~\citep{han2506vision, peng2022beit}.

%% file: sections/2_related.tex
\section{Related Work}
\label{sec:related}

% Paragraph 1: Unified Multimodal Models - Background
Unified models (UMs) process and create content across multiple modalities in one architecture.
Early approaches enable multimodal generation by extending pretrained LLMs with discrete visual tokens~\citep{team2024chameleon, wang2024emu3, kou2024orthus, xie2025show, zhou2024transfusion, chern2024anole, sun2023emu, tong2025metamorph}.
Representative works include Chameleon~\citep{team2024chameleon} and Emu3~\citep{wang2024emu3} that employ unified transformers for next-token prediction across modalities, Show-o~\citep{xie2025show} and Transfusion~\citep{zhou2024transfusion} that combine autoregressive modeling with diffusion.
More recent efforts explore unified architectures with improved training strategies~\citep{chen2025janus, ma2025janusflow, deng2025emerging, cui2025emu3} and interleaved image-text generation capabilities~\citep{chern2024anole, ye2025echo, hao2025uni}.
Despite these advances, most UMs primarily focus on visual generation tasks such as text-to-image synthesis and image editing~\citep{wu2025qwen, wan2025, flux2024}.
We discuss more related work in Appendix~\ref{app:related}.

%% file: sections/5_conclusion.tex
\section{Conclusion}
\label{sec:conclusion}
We introduced \MODEL, a latent-space unified model in which generated visual tokens remain directly interpretable by the model for model-internal cross-modal reasoning.
By combining MBAQ, MoME, and an optional pixel decoder, \MODEL avoids pixel-space mediation during reasoning while still supporting visualization.
Empirically, \MODEL preserves strong visual understanding, outperforms unified baselines on GenEval and GenEval2, reaches state-of-the-art performance on Visual Spatial Planning, and shows preliminary action-conditioned world-modeling capability.

Several limitations remain.
At the model level, \MODEL is trained at a modest scale and currently supports fixed-resolution generation.
Moreover, although the MBAQ objective transfers across VLM families, the learned quantizer is backbone-specific and must be retrained for each target VLM.
At the application level, self-reward for visual generation has so far been validated only with compositional verification questions, and fully latent long-horizon rollout in world modeling can be further strengthened for tasks requiring precise spatial fidelity.
Future work should therefore scale the model and data, support variable-resolution and longer-context generation, design holistic rewards for open-ended generation, and further develop fully latent temporal prediction for planning.

%% file: sections/apdx5_related.tex
\section{More Related Work}
\label{app:related}

% Paragraph 2: Visual Representation for UMs - Core Challenge
Building effective UMs requires visual representations for both understanding and generation.
Prior work argues that these two tasks require fundamentally different features—semantic features for understanding~\citep{radford2021learning, tschannen2025siglip} versus pixel details for generation~\citep{van2017neural, esser2021taming, kingma2013auto}—leading to approaches that adopt separate visual encoders~\citep{wu2025janus, chen2025janus, huang2025illume+, hao2025uni, li2025unifork, yan2025can} or dual vocabularies~\citep{song2025dualtoken, li2025manzano, qu2025tokenflow}.
Other works pursue unified tokenizers with shared representations~\citep{ma2025unitok, wu2024vila, liu2025tuna, wu2025harmonizing, peng2022beit, sun2024autoregressive, ai2025ming, li2025onecat, fan2025prism, zhang2025unilip, lin2025toklip}, or explore semantic features directly for generation~\citep{chen2025diffusion, zheng2025diffusion, du2025vqrae, han2506vision, geng2025x}.

% Paragraph 3: Cross-Modal Reasoning
Most UMs focus on single-turn generation or understanding, but a growing body of work investigates interleaved reasoning.
Visual Chain-of-Thought extends text-based CoT~\citep{wei2022cot, kojima2022zeroshot} to multimodal settings.
Some approaches generate explicit intermediate visual steps~\citep{hu2024visualsketchpad, xu2024llavacot, chen2026show}, while others reason implicitly in latent spaces~\citep{li2025lvr, yang2025machine, li2025imagine, hao2024training, deng2024explicit}.
Most relevant to our work are ThinkMorph~\citep{gu2025thinkmorph}, Uni-CoT~\citep{qin2025uni}, and UniCorn~\citep{han2026unicorn}, which study interleaved text-image reasoning traces.
These approaches, however, rely on pixel-space mediation between visual understanding and generation~\citep{gu2025thinkmorph, qin2025uni}, where visual features must be decoded to pixels and re-encoded for understanding, introducing codec bias and cross-modal misalignment~\citep{yi2024bridge, fan2025prism}.

% Paragraph 4: World Modeling
World models predict future environmental states conditioned on past observations and actions~\citep{ha2018world,lecun2022path,agarwal2025cosmos,xing2025critiques}, a capability that naturally aligns with the unified understanding-and-generation objective of UMs.
Recent unified models have begun to embrace this connection: Emu3.5~\citep{cui2025emu3} demonstrates that next-token prediction on interleaved multimodal sequences yields generalizable world-modeling abilities, and Transfusion-RAE~\citep{tong2026beyond} finds that unified multimodal pretraining on action-conditioned video naturally gives rise to world modeling. \MODEL adopts the same perspective, predicting future visual states as semantic tokens in the shared latent space and treating world modeling as an instance of cross-modal reasoning.

%% file: sections/apdx1_training_detail.tex
\section{Training Details}
\label{app:pretraining}

\subsection{Experimental Setup}
\label{app:exp_setup}
% \begin{figure*}[t]
%     \centering
%     \includegraphics[width=\textwidth]{fig/demo_lighter.pdf}
%     \caption{\textbf{Text-to-image gallery of \MODEL.} The last column highlights \MODEL's text rendering capability, which emerges as visual and language tokens share a unified semantic space, enabling legible in-image text.}
%     \label{fig:demo}
%     \vspace{-6pt}
% \end{figure*}
During the pre-training of \MODEL{}\textsubscript{Base}, the visual generation resolution is fixed to $448 \times 448$, represented by 256 tokens. The quantizer contains 101M trainable parameters, with each latent vector chunked into 8 parts and a codebook of 2048 codes per part~\citep{ma2025unitok}. The visual generation branch comprises 3633M parameters, and the autoregressive head is a 3-layer transformer with 283M parameters. For the decoupled pixel decoder, we fine-tune the conditioning branch (990M parameters) of MMDiT from Stable Diffusion 3.5 Medium~\citep{esser2024scaling}.

\subsection{MBAQ Tokenizer Training}
\label{app:mbaq_training}
The MBAQ quantizer follows a compress-quantize-expand pipeline.
A two-layer MLP with GELU activation projects the input CLIP features from $d{=}4096$ to a bottleneck dimension of $d_e{=}256$.
The projected features are quantized by a multi-codebook quantizer: the 256-dimensional vector is evenly partitioned into $C{=}8$ chunks of 32 dimensions each, and each chunk is independently quantized using a separate codebook of size $K{=}2048$, yielding an effective vocabulary of $K^C = 2048^8$ from only $K \times C = 16{,}384$ total codebook entries.
A second two-layer MLP with GELU activation then projects the de-quantized features back to the LLM hidden size of 2560 for downstream language model consumption.
Codebook embeddings are initialized uniformly in $[-1/K,\, 1/K]$.

The full training objective is:
\begin{equation}
    \mathcal{L}(\phi) = \mathcal{L}_{\text{KL}} + \lambda\, \mathcal{L}_{\text{MCQ}}, \quad \lambda = 1.0,
\end{equation}
where the KL divergence loss $\mathcal{L}_{\text{KL}}$ is computed only over answer token positions (Section~\ref{sec:unified_repr}).
The MCQ commitment loss $\mathcal{L}_{\text{MCQ}}$ is averaged across all $C$ codebooks:
\begin{equation}
    \mathcal{L}_{\text{MCQ}} = \frac{1}{C} \sum_{c=1}^{C}
    \Bigl[
        \underbrace{\|\mathbf{z}_q^c - \mathrm{sg}[\mathbf{z}^c]\|_2^2
        + \beta\,\|\mathrm{sg}[\mathbf{z}_q^c] - \mathbf{z}^c\|_2^2}_{\text{codebook + commitment loss}}
        \;-\; \underbrace{\alpha\, H(\bar{p}^c)}_{\text{entropy regularizer}}
    \Bigr],
\end{equation}
where $\mathrm{sg}[\cdot]$ denotes the stop-gradient operator, $\beta{=}0.25$ is the commitment cost, and $\alpha{=}0.1$ weights the entropy regularizer.
$H(\bar{p}^c) = -\sum_k \bar{p}^c_k \log \bar{p}^c_k$ is the entropy of the soft code assignment distribution $\bar{p}^c = \mathrm{mean}_{b,\ell}\,\mathrm{softmax}(-\mathbf{d}^c_{b,\ell})$, where $\mathbf{d}^c_{b,\ell}$ contains the squared distances to all codebook entries for position $(b, \ell)$ in codebook $c$.
Maximizing this entropy regularizes each codebook toward uniform code usage.

Gradient propagation through the quantization step uses the straight-through estimator (STE)~\citep{bengio2013estimating}.
Codebook utilization is tracked via EMA with decay $\gamma{=}0.99$.
Codes whose EMA usage falls below $0.03/K$ are identified as dead codes and restarted: at most 64 dead codes are reset per step by replacing their embeddings with randomly sampled inputs from the current batch.

\subsection{MoME Pretraining}
\label{app:mome_training}
The MoME generation branch is constructed by adding dedicated \emph{vision} counterparts to every transformer layer of InternVL3.5-4B, following the Mixture-of-Transformers (MoT) design~\citep{deng2025emerging}.
Specifically, for each Qwen3 decoder layer, we duplicate the Q/K/V/O projection matrices and Q/K normalization layers into vision-specific copies (\texttt{q/k/v/o\_proj\_vision}, \texttt{q/k\_norm\_vision}), and add a separate vision MLP (\texttt{mlp\_vision}).
All vision-specific parameters are initialized from the corresponding pretrained understanding parameters.
At each forward pass, a \texttt{vision\_token\_mask} routes each token to the appropriate branch: visual generation tokens are processed by the \texttt{\_vision} parameters, while text and understanding tokens use the original frozen parameters.
Both branches share the same key--value attention computation, enabling cross-modal interaction through shared self-attention. A two-layer MLP projector maps the de-quantized MBAQ features into the LLM hidden space before being concatenated with text embeddings.
The visual understanding branch (vision encoder and original LLM parameters) is fully frozen throughout pretraining.

The AR vision head is a 3-layer causal Transformer decoder with hidden size 2560, 32 attention heads, MLP ratio 4.0, RMSNorm, and 1D rotary position embeddings.
It takes the MoME hidden state at each visual token position as a prefix and autoregressively predicts $C{=}8$ codebook indices in sequence, implementing the MC-NTP objective (Eq.~\ref{eq:mcntp}).

The pretraining objective is cross-entropy over the predicted codebook indices as defined in Eq.~\ref{eq:mcntp}.
We apply classifier-free guidance (CFG) training by randomly dropping the text conditioning with probability 0.1.
The model is trained on 32M text-image pairs from the BLIP3o dataset~\citep{chen2025blip3o} (Short-Caption, Long-Caption, and JourneyDB subsets) at a fixed resolution of $448{\times}448$ pixels (256 tokens per image) with a maximum sequence length of 512 (including the text prompt token length).
MoME pretraining takes approximately 4 days on 64$\times$NVIDIA H100 GPUs.

\begin{figure}[!th]
    \centering
    \includegraphics[width=1\textwidth]{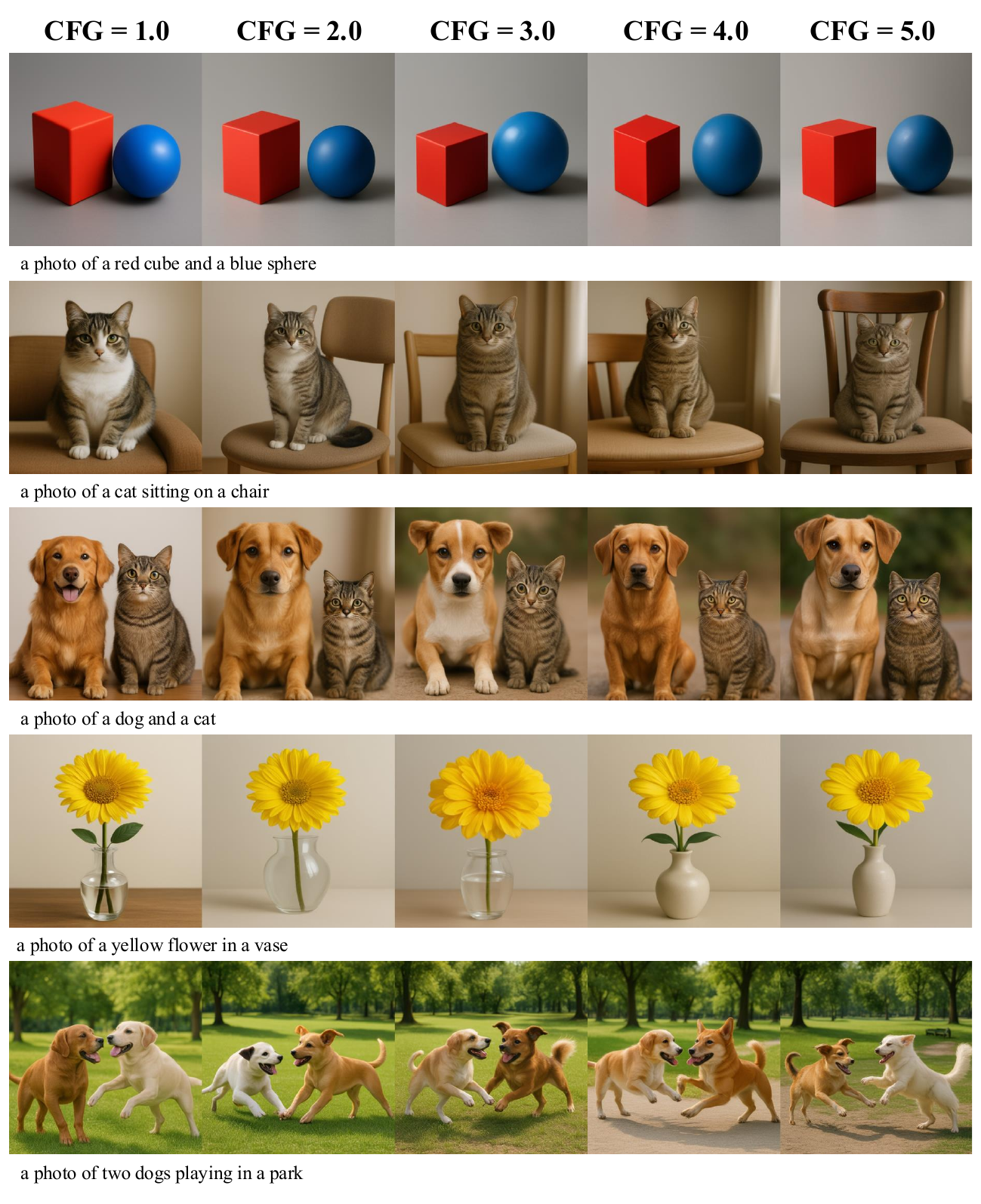}
    \caption{Generation results under different CFG scales ($1.0$ to $5.0$)
 for identical prompts. \MODEL produces coherent images even at CFG $=1.0$.}
    \label{fig:cfg_comparison}
\end{figure}

At inference time, visual tokens are sampled autoregressively using nucleus sampling with top-$k{=}32$, top-$p{=}0.95$, and temperature $\tau{=}0.5$.
For standard benchmark evaluations, classifier-free guidance (CFG) is applied with scale $s{=}3.0$, where the unconditional prediction is obtained by dropping the text condition.
Benefiting from the semantic alignment induced by MBAQ, \MODEL also supports CFG-free generation ($s{=}1.0$): since the generative visual tokens share the same semantic space as language, the model produces semantically coherent, instruction-following images even without CFG, a capability not available to models based on pixel-reconstruction features.
Figure~\ref{fig:cfg_comparison} presents side-by-side comparisons of images generated with different CFG scales under identical prompts.

\subsection{Pixel Decoder Training}
\label{app:decoder_training}
The pixel decoder is built upon the MMDiTX architecture of Stable Diffusion 3.5 Medium~\citep{esser2024scaling}, which consists of 24 dual-stream transformer layers with a patch size of 2 operating on 16-channel VAE latents.
We replace the original text conditioning module with a learned linear projector that maps the 256-dimensional MBAQ de-quantized embeddings to the MMDiT context dimension.
The trainable parameters consist of the \texttt{context\_embedder} and all \texttt{context\_block} parameters (the conditioning stream of each transformer layer), totaling approximately 990M parameters.
All remaining parameters, including the image stream (\texttt{x\_embedder}, \texttt{x\_block}), the VAE, and the MBAQ quantizer are fully frozen throughout training.

The decoder is trained with a flow matching objective~\citep{esser2024scaling}.
Timesteps are sampled via logit-normal distribution with mean $\mu{=}0.0$, standard deviation $\sigma{=}1.0$, and mode scale 1.29, following the SD3 training recipe.
The training loss is a weighted mean squared error between the predicted and target latents:
\begin{equation}
    \mathcal{L}_{\text{diff}}(\eta) = \mathbb{E}_{u, \boldsymbol{\epsilon}, \mathbf{x}_0}
    \left[ w(\sigma_u) \, \| \hat{\mathbf{x}}_0 - \mathbf{x}_0 \|^2 \right],
\end{equation}
where $\hat{\mathbf{x}}_0$ is the model's predicted clean latent, $\sigma_u$ is the noise level at timestep $u$, and $w(\sigma_u)$ is the logit-normal loss weight.

The decoder is pretrained on the same 32M text-image pairs from BLIP3o~\citep{chen2025blip3o} used for MoME pretraining, with per-GPU batch size 45.
This stage takes approximately 2 days on 32$\times$NVIDIA H100 GPUs.

At inference time, the pixel decoder uses a Flow Matching Euler discrete scheduler~\citep{esser2024scaling} with $N{=}25$ denoising steps at a fixed resolution of $448{\times}448$.
The MBAQ de-quantized features $\tilde{\mathbf{V}}$ serve as the sole conditioning signal, with classifier-free guidance scale set to $s{=}1.0$ (i.e., no unconditional guidance branch).
This is possible because $\tilde{\mathbf{V}}$ already carries rich semantic content from the MBAQ-aligned latent space, providing sufficiently strong conditioning. This phenomenon is also reported by X-Omni which also conducts autoregressive generation over semantic tokens~\citep{geng2025x}.

%% file: sections/apdx2_self_reflection.tex
\section{Additional Analysis of Visual Generation with Self-Reflection}
\label{app:self_reflection_analysis}
\subsection{GRPO setup}
\begin{figure}[h]
    \centering
    \includegraphics[width=1\textwidth]{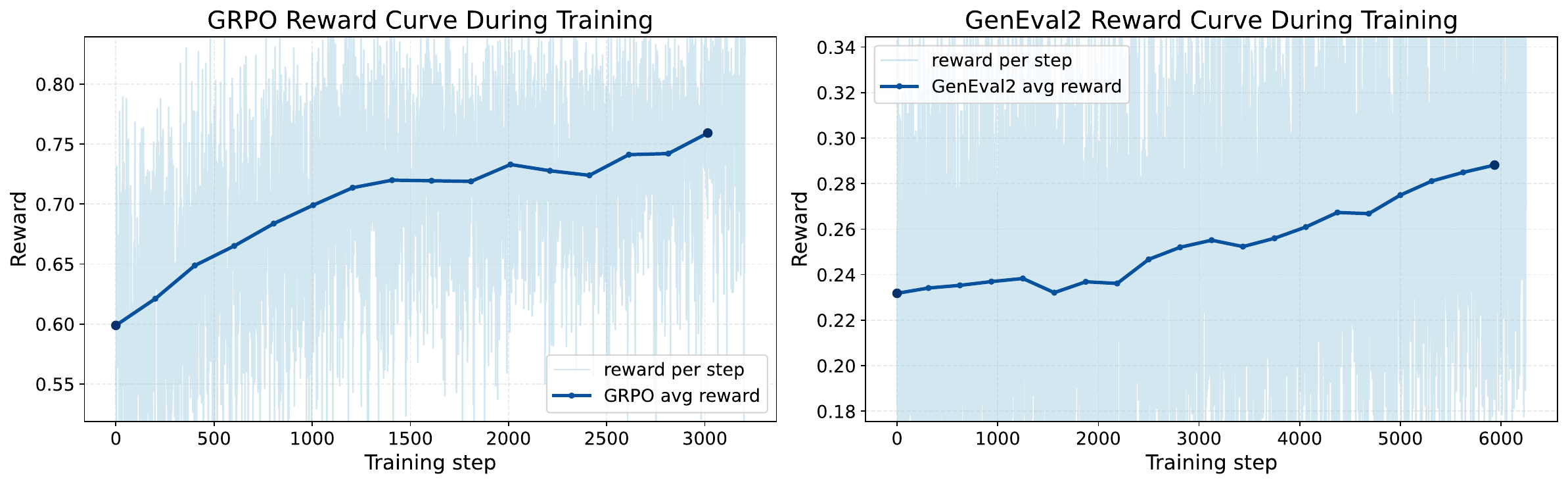}
    \caption{Reward curve during GRPO training for self-reflection visual generation.}
    \label{fig:grpo_reward}
\end{figure}
For visual generation with self-reflection, we optimize \MODEL{}\textsubscript{Base} with GRPO using prompt-wise grouped rollouts.
Each prompt produces $G{=}16$ samples, and advantages are normalized within each prompt group.
We use PPO-style clipping with clip parameter $0.1$ and KL regularization weight $\beta{=}0.005$.
Rollouts are generated without classifier-free guidance, using guidance scale $1.0$, sampling temperature $0.9$, top-$p{=}0.95$.

The trainable parameters are restricted to the visual-generation pathway introduced by MoME, including the vision-specific transformer parameters, the visual projector, and the autoregressive vision head.
This restriction keeps post-training focused on improving generation behavior while preserving the pretrained understanding backbone that supplies the self-reward signal.
Concretely, the understanding branch $\psi$ is frozen throughout self-reward RL and is used only as the evaluator, while the visual-generation pathway $\theta$ is updated as the policy. The pixel decoder is not used for reward computation; rewards are computed directly from the generated semantic visual features, avoiding an additional decode-and-detect loop during training.

\subsection{Prompt decomposition into verification questions}
\label{app:prompt_decomposition}
The self-reward is computed by decomposing each generation prompt into a set of atomic visual checks and then asking the understanding branch to answer them as multiple-choice questions.
The decomposition covers the semantic factors evaluated by GenEval-style benchmarks, including object existence, object count, color, material, pattern, and pairwise spatial relations.
For each prompt, we derive the ground-truth answer to each check directly from the prompt specification. Object existence questions are created for mentioned objects, count questions are created when cardinality is specified, attribute questions bind colors/materials/patterns to their target objects, and relation questions check pairwise spatial constraints. Each check is converted into a multiple-choice question with alphabet-prefixed options, such as \texttt{A. 0}, \texttt{B. 1}. The following is an example:
\begin{center}
\setlength{\fboxsep}{8pt}
\fcolorbox{black!30}{black!3}{%
\begin{minipage}{0.94\linewidth}
\small

\vspace{3pt}
\textbf{Prompt:} ``five black teddy bears''

\vspace{4pt}
\textbf{Q1 (count).}
Answer the following question with the given option only.
How many valid teddy bears are there in the image?
Valid means the object is clearly visible, complete and not partially occluded by other objects and contains no artifacts.
\newline
\textit{Options:}
A: 0, B: 1, C: 2, D: 3, E: 4, F: 5, G: 6, H: 7, I: 8, J: 9, K: 10.
\newline
\textit{Ground-truth option:} F (5).

\vspace{4pt}
\textbf{Q2 (color).}
Answer the following question with the given option only.
What color is the teddy bear in the image?
\newline
\textit{Options:}
A: red, B: orange, C: yellow, D: green, E: blue, F: purple, G: pink, H: brown, I: black, J: white.
\newline
\textit{Ground-truth option:} I (black).

\vspace{4pt}
\textbf{Q3 (existence).}
Answer the following question with the given option only.
Does a/an teddy bear exist in the image?
Answer yes only when the object is clearly visible, complete and not partially occluded by other objects and contains no artifacts.
\newline
\textit{Options:} A: Yes, B: No.
\newline
\textit{Ground-truth option:} A (Yes).

\vspace{5pt}
\end{minipage}%
}
\end{center}

During rollout, the generated latent visual sequence is first mapped back to semantic visual features.
The understanding branch then evaluates these features with each verification question without explicitly decoding a textual answer.
Instead, at the answer position, we directly extract the logits of all candidate option tokens (e.g., A/B/C/...) and apply a softmax over this restricted option set.
Specifically, for a verification question $\nu_m$ with candidate option-token logits $\{\rho_{m,k}\}_k$ and ground-truth option index $k^\star_m$, we compute the per-question score as $p_m = \frac{\exp(\rho_{m,k^\star_m}/\tau)}{\sum_k \exp(\rho_{m,k}/\tau)}$. This uses the first option-token logits only, so reward evaluation does not require free-form answer decoding.

If a prompt is decomposed into $M$ verification questions and the resulting ground-truth option probabilities are $\{p_m\}_{m=1}^{M}$, we define the self-reward as
\begin{equation}
    \mathcal{R} = \prod_{m=1}^{M} p_m.
\end{equation}
This multiplicative design is intentionally strict: a sample receives a high reward only when it satisfies all required constraints simultaneously, rather than only a subset of them.
As a result, the reward favors globally correct generations with consistent object counts, attributes, and relations.

\subsection{Qualitative Comparison Before and After RL}
\begin{figure}[h]
    \centering
    \includegraphics[width=1\textwidth]{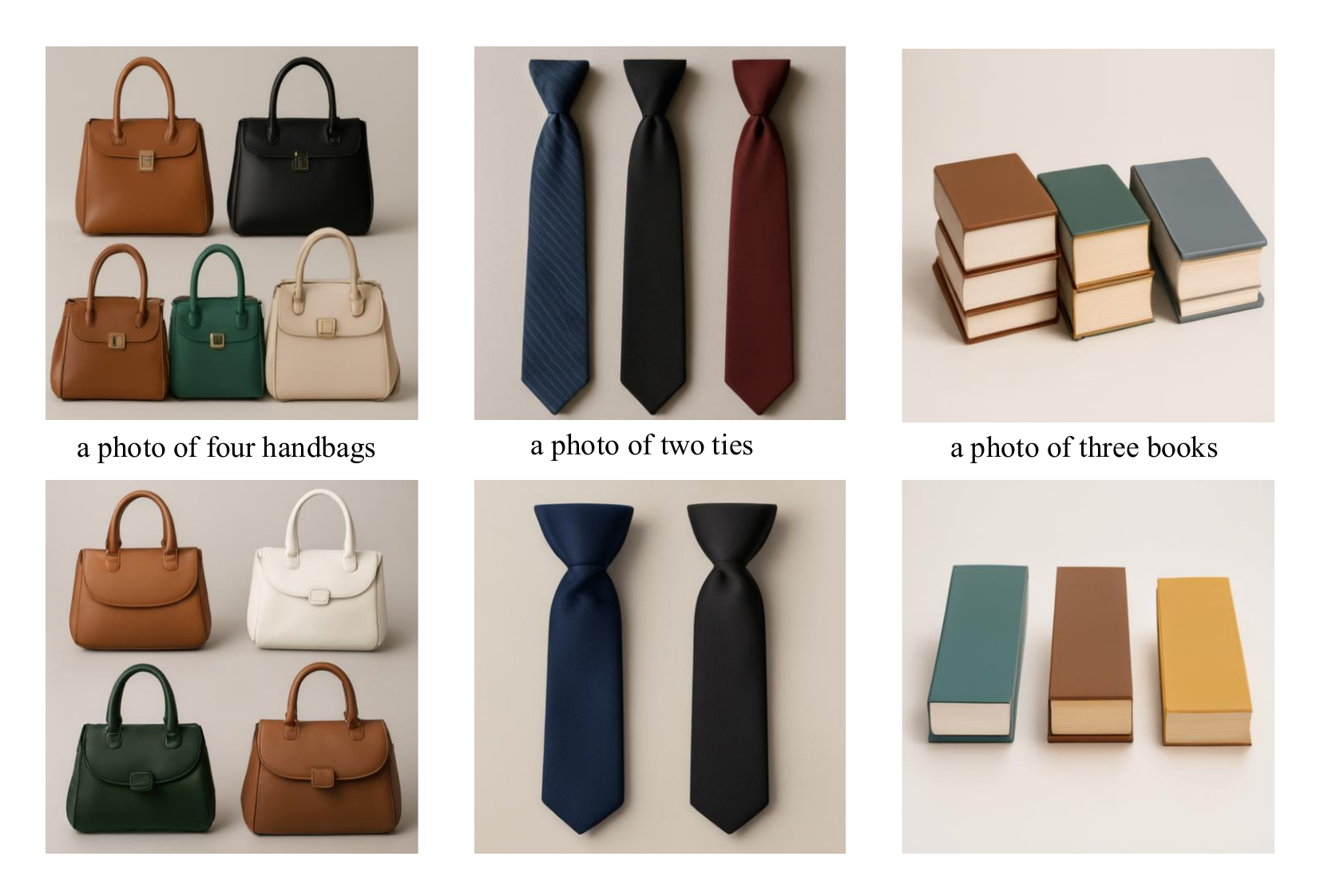}
    \caption{Qualitative comparison before and after self-reward GRPO. Top:
  \MODEL{}\textsubscript{Base} before GRPO. Bottom: the model after self-reward GRPO. The post-trained model follows counting prompts more
  accurately.}
    \label{fig:grpo_effect}
\end{figure}
Figure~\ref{fig:grpo_reward} shows that the reward increases steadily during GRPO training.
Figure~\ref{fig:grpo_effect} further shows the qualitative effect of self-reward GRPO.
Compared with \MODEL{}\textsubscript{Base}, the post-trained model follows compositional constraints more accurately, especially for counting, attribute binding, and spatial relations.
This result is consistent with the design of the reward, which directly encourages semantic correctness rather than low-level visual fidelity.

\subsection{Failure Cases of Detection-Based Metrics}
\label{sec:failure_detection_metric}
\begin{figure}[h]
    \centering
    \includegraphics[width=1\textwidth]{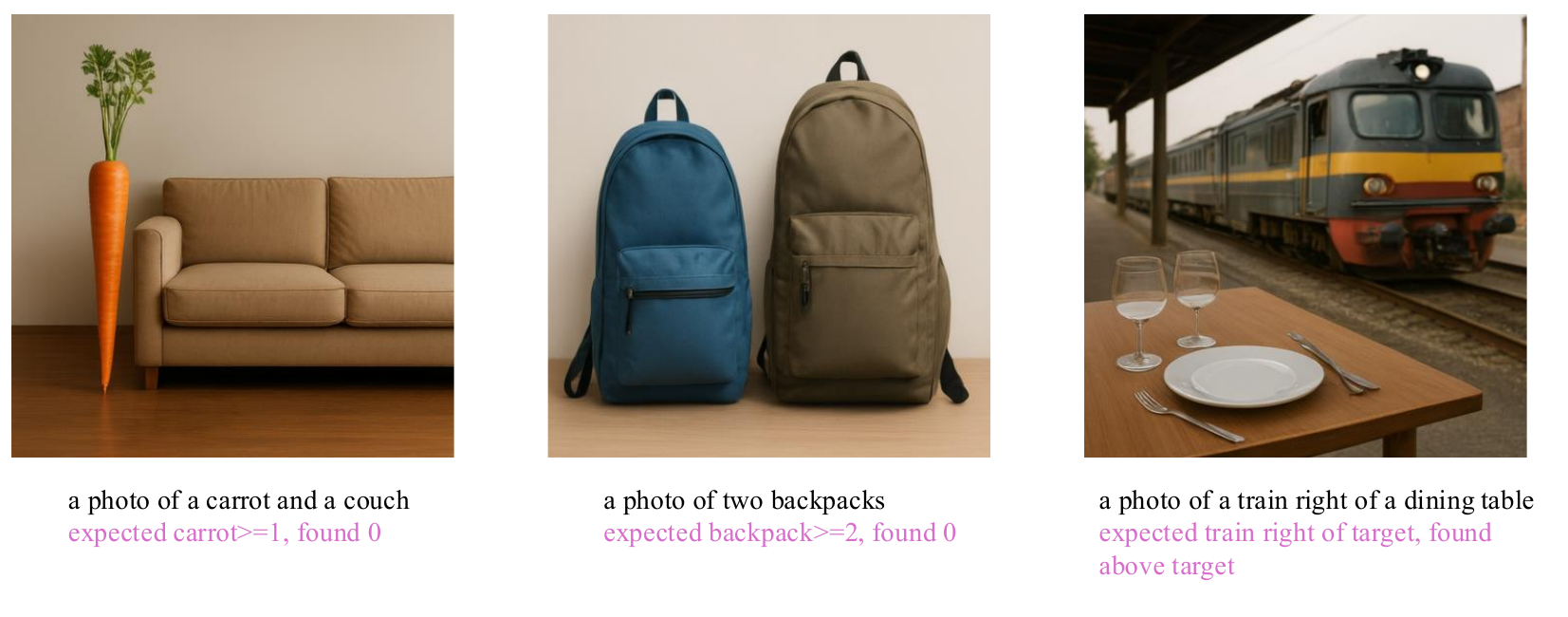}
    \caption{Examples where the generated images are qualitatively correct, but the GenEval detection-based metrics mark them as failures.}
    \label{fig:detector_failure}
\end{figure}
Figure~\ref{fig:detector_failure} shows representative examples in which detection-based metrics incorrectly judge the generated samples as failures.
In these cases, the generated images are qualitatively correct, but the GenEval detection-based metric still marks them as failures. This indicates that detection-based metrics are not always reliable for evaluating semantic instruction following. This problem is especially visible for dense layouts, small objects, and visually unusual but valid renderings.
As a result, a detection-based metric may reflect detector brittleness rather than the true semantic correctness of the generated image.

This is also why we introduce GenEval2 with a VLM-as-a-judge evaluator in the
main text.
Compared with detection-based scoring, a VLM judge is better aligned with the semantic content of the prompt and is less brittle to visual variation.

%% file: sections/apdx3_fine_vsp_data.tex
\section{Fine-Grained VSP Data Construction}
\label{app:vsp_data}
We construct the fine-grained VSP training set by procedurally generating deterministic FrozenLake mazes and converting each solvable maze into an interleaved action-image trajectory.
The benchmark contains four difficulty levels corresponding to $3\times3$ to $6\times6$ grids, and we generate mazes at the corresponding scales.
For a maze of size $n\times n$, we first sample two distinct cells as the start $S$ and goal $G$.
Every remaining cell is independently assigned to either frozen land $F$ (traversable) or a hole $H$ (blocked), yielding a random layout $m \in \{S,F,H,G\}^{n\times n}$.

\paragraph{Solvability and path filtering.}
We treat each non-hole cell as a node in a 4-neighbor grid graph and compute a shortest path from $S$ to $G$ with breadth-first search.
Layouts with no valid path are discarded.
We further retain only mazes whose shortest-path length lies in a predefined interval, which removes trivial instances and controls the planning horizon.
To reduce repeated supervision on identical maps, we de-duplicate mazes by their flattened grid string and keep only one copy of each unique layout.
Each retained example therefore corresponds to one unique solvable maze and one expert shortest-path trajectory $s_0, s_1, \ldots, s_T$.

\paragraph{Step-wise rendering.}
Given the trajectory, we derive an action sequence $a_t \in \{U, D, L, R\}$ from consecutive state pairs $(s_{t-1}, s_t)$.
We then render every intermediate state directly from the deterministic FrozenLake simulator.
For step $t$, the rendered image shows the agent at state $s_t$ and overlays all previously visited cells in red; when the agent reaches the goal, the goal cell is additionally highlighted in green.
The rendered images are resized to a fixed resolution determined by the maze size, which yields visually consistent supervision across difficulty levels.

\paragraph{Interleaved supervision format.}
We finally convert each trajectory into the fine-grained format shown in Figure~\ref{fig:vsp}(b).
The user prompt contains only the initial image $I_0$, while the target response is an alternating sequence of actions and next-state images:
\[
a_1~\texttt{<img>}~a_2~\texttt{<img>}~\cdots~a_T~\texttt{<img>}~\texttt{Done!},
\]
where the $t$-th \texttt{<img>} slot is supervised by the rendered image $I_t$ after executing action $a_t$.
This format forces the model to alternate between textual action prediction and visual state updating at every step, rather than solving the maze with a single coarse plan.

%% file: sections/apdx4_decoding_ref.tex
\section{Improving Decoding Consistency in Long-Horizon Prediction}
\label{app:wm}
\begin{figure}[h]
    \centering
    \includegraphics[width=0.95\textwidth]{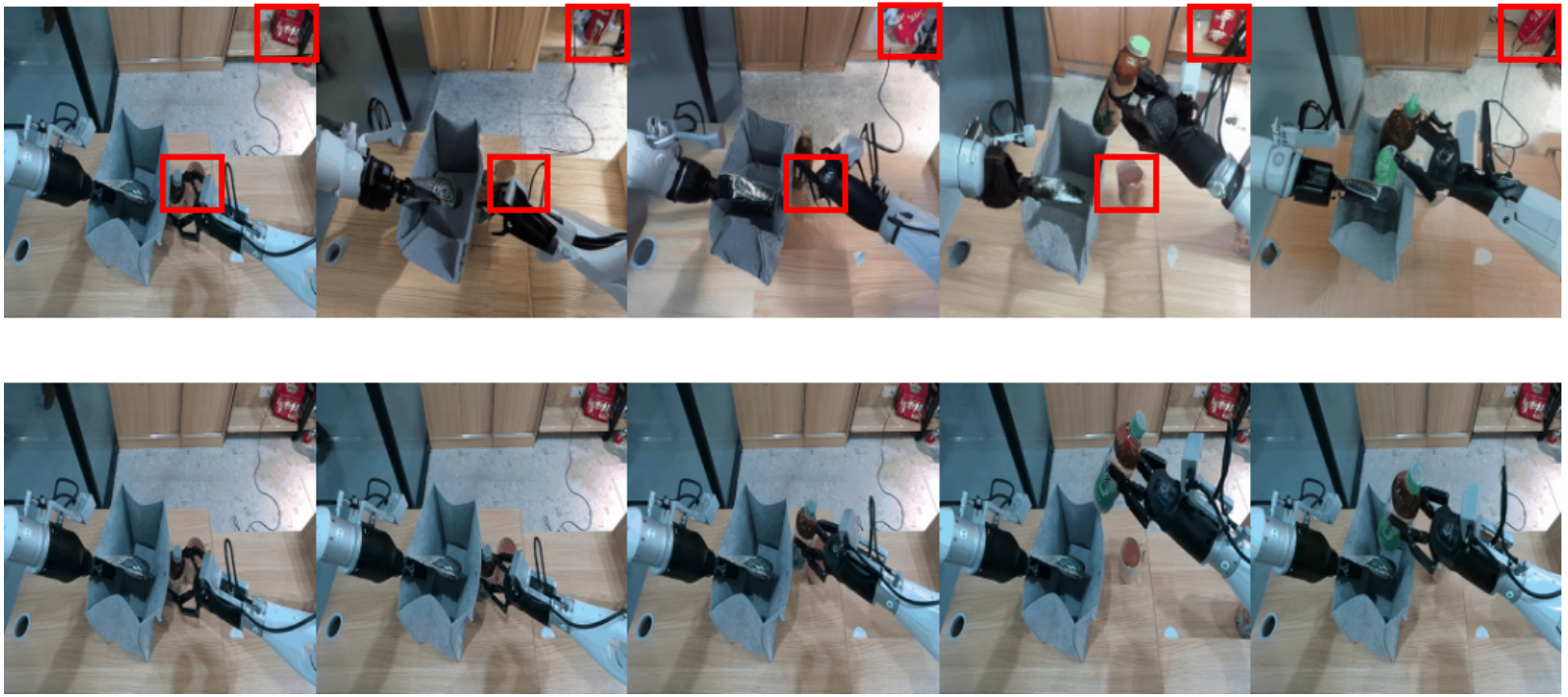}
    \caption{Top: independent keyframe decoding with the original decoupled decoder. Bottom: reference-conditioned decoding with the first-frame VAE latent. The bottom row shows much better cross-frame consistency.}
    \label{fig:decoder_ref}
\end{figure}
For long-horizon prediction tasks, such as robot-motion forecasting or video keyframe prediction, semantic correctness alone is not sufficient.
The decoded frames should also remain visually consistent across time, so that object identity, scene layout, and appearance do not drift from one step to the next.
In practice, this is a challenge for a decoupled decoder, because decoding from semantic tokens alone may recover the correct high-level content while still introducing frame-to-frame inconsistency in low-level appearance.
Figure~\ref{fig:decoder_ref} shows that adding a first-frame reference during decoding leads to much more consistent keyframe predictions.

To address this issue, we augment the decoder with an additional reference-image condition.
In addition to the quantized semantic tokens of the target frame, we feed the decoder with the VAE latent of a reference frame, chosen as the first frame of the predicted sequence.
Intuitively, the semantic tokens specify \emph{what} should be rendered at the current step, while the reference latent provides a stable visual anchor for \emph{how} the scene should look across the sequence.
This design preserves the flexibility of semantic prediction while improving temporal consistency during decoding.

We find that this reference-conditioned decoder substantially improves the consistency of decoded keyframes in long-horizon settings.
In particular, it is better at maintaining stable appearance over extended prediction horizons, which is important for applications such as world models and long-term visual planning that requires pixel space information.
These results suggest that when decoded frames are used as intermediate states rather than standalone images, incorporating an explicit reference signal is an effective way to reduce visual drift in rendered pixels.

%% file: sections/apdx6_zero-shot_language.tex
\section{Zero-shot language-conditioned navigation}
\label{app:wm_lang}
\begin{figure}[h]
    \centering
    \includegraphics[width=0.95\textwidth]{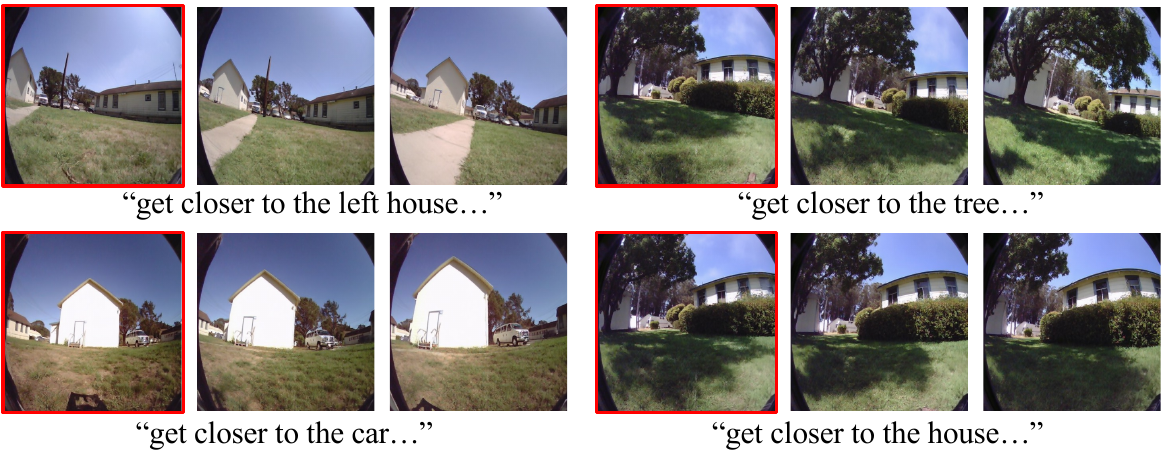}
    \caption{Without language supervision during fine-tuning, \MODEL{}\textsubscript{WM} translates free-form textual instructions (e.g., “get closer to the left house”) into accurate future visual states, enabled by the inherited language-vision alignment.}
    \label{fig:zero_shot_lang}
\end{figure}
As an emergent capability, we observe that \MODEL{}\textsubscript{WM} can perform zero-shot language-conditioned navigation. Although the model is fine-tuned exclusively on standard action tokens without any free-form language supervision for navigation tasks, it can generalize to natural language instructions. 

As illustrated in Figure~\ref{fig:zero_shot_lang}, given an unconstrained textual prompt such as ``get closer to the left house,'' the model leverages the rich language-vision alignment inherited from \MODEL{}\textsubscript{Base} to predict temporally and semantically consistent future visual states. This highlights the robustness and flexibility of the unified semantic latent space in interpreting and grounding open-ended textual conditions into physical-world dynamics.

%% file: sections/apdx7_additional_exp.tex
\section{Additional Experiments}
\label{app:additional_exp}

\subsection{Data-Matched Comparison with ThinkMorph on VSP}
\label{app:vsp_data_matched}
To isolate the effect of the latent representation from task-specific supervision, we retrain ThinkMorph~\citep{gu2025thinkmorph} on exactly the same fine-grained VSP dataset and alternating action-image reasoning format as \MODEL{}\textsubscript{Vis-Plan} (Appendix~\ref{app:vsp_data}).
ThinkMorph is initialized from BAGEL-14B~\citep{deng2025emerging}, a substantially larger backbone than that of \MODEL.
Because it relies on pixel-space mediation, each intermediate visual state requires processing ViT features, VAE features, and noisy VAE features, resulting in roughly three times the visual representation overhead.
Consequently, on the same cross-modal reasoning trajectories, ThinkMorph produces sequences approximately 4--5$\times$ longer than \MODEL, making each training step several times more computationally expensive.

\begin{table}[h]
    \centering
    \caption{Data-matched comparison on VSP~\citep{wu2024vsp}. Both models are trained on the same fine-grained VSP data with the same alternating action-image reasoning format.}
    \label{tab:vsp_data_matched}
    \begin{tabular}{@{}lccccc@{}}
        \toprule
        Model & L3 & L4 & L5 & L6 & Avg. \\
        \midrule
        ThinkMorph~\citep{gu2025thinkmorph} & 0.92 & 0.83 & 0.67 & 0.43 & 0.71 \\
        \MODEL{}\textsubscript{Vis-Plan} & 1.00 & 1.00 & 1.00 & 0.97 & 0.99 \\
        \bottomrule
    \end{tabular}
\end{table}

As shown in Table~\ref{tab:vsp_data_matched}, ThinkMorph still underperforms \MODEL despite its larger backbone and a training compute budget substantially exceeding that used for \MODEL.
Inspection of the evaluation rollouts shows that its intermediate images become progressively noisier over later reasoning steps, indicating stronger error accumulation from repeated pixel-space decoding and re-encoding.
This issue is particularly pronounced on higher-level VSP cases, which require longer reasoning trajectories and therefore amplify accumulated visual errors.

\subsection{Out-of-Distribution Generalization on VSP}
\label{app:vsp_ood}
To evaluate out-of-distribution generalization, we test the same \MODEL{}\textsubscript{Vis-Plan} checkpoint without any additional fine-tuning under five controlled shifts:
(1) \emph{unseen object appearances}, where the hole, goal, and agent are drawn with entirely different artwork;
(2) \emph{color shift and image degradation}, including hue rotation, contrast scaling, Gaussian noise, and JPEG compression;
(3) \emph{grayscale rendering}, which removes the color cue separating holes from the goal so that they must be distinguished by shape alone;
(4) \emph{larger grid}, extending the maze from $6\times6$ to $7\times7$ at a fixed hole density; and
(5) \emph{confusable distractors}, where objects visually similar to holes and to the goal are placed on safe cells.
We generate 100 new mazes for each setting.
In addition to the task success rate, we report the step-wise optimal-action rate, defined as the fraction of predicted actions that strictly reduce the shortest-path distance to the goal.

\begin{table}[h]
    \centering
    \caption{Out-of-distribution generalization of \MODEL{}\textsubscript{Vis-Plan} on VSP without additional fine-tuning.}
    \label{tab:vsp_ood}
    \begin{tabular}{@{}lcc@{}}
        \toprule
        Setting & Success rate & Step-wise optimal-action rate \\
        \midrule
        In-distribution ($6\times6$) & 0.97 & 0.99 \\
        Unseen object appearances ($6\times6$) & 0.61 & 0.89 \\
        Color shift and degradation ($6\times6$) & 0.39 & 0.97 \\
        Grayscale rendering ($6\times6$) & 0.12 & 0.91 \\
        Larger grid ($7\times7$) & 0.41 & 0.87 \\
        Confusable distractors ($6\times6$) & 0.59 & 0.85 \\
        \bottomrule
    \end{tabular}
\end{table}

As shown in Table~\ref{tab:vsp_ood}, the success rate decreases under all out-of-distribution settings, whereas the step-wise optimal-action rate remains relatively high.
Our error analysis suggests that much of the drop in end-to-end success is caused by failures in termination detection rather than in spatial planning: the model often continues to select actions that correctly move it toward the goal, but fails to recognize that the goal has been reached because the out-of-distribution transformations alter the visual pattern used to identify the goal state.
These results indicate that the learned policy retains substantial local planning capability under visual shifts, while its goal-state recognition remains sensitive to appearance changes.

\subsection{Cross-VLM Generalization of MBAQ}
\label{app:mbaq_cross_vlm}
MBAQ is not intrinsically tied to InternVL3.5: its objective only requires a VLM that consumes continuous visual features and produces language-output distributions, so the same formulation can be applied to other VLMs.
To verify this, we repeat MBAQ training and evaluation with Qwen3-VL-2B-Instruct~\citep{Qwen3-VL}.

\begin{table}[h]
    \centering
    \caption{MBAQ applied to Qwen3-VL-2B-Instruct. MBAQ discretizes the continuous features of the Qwen3-VL vision encoder while preserving most of the original visual understanding capability.}
    \label{tab:mbaq_cross_vlm}
    \begin{tabular}{@{}lccccc@{}}
        \toprule
        Model & MME-P & POPE & SEED & MMBench & MMMU\textsubscript{val} \\
        \midrule
        Qwen3-VL-2B (continuous) & 1493 & 88.6 & 75.1 & 78.0 & 44.9 \\
        Qwen3-VL-2B (discretized via MBAQ) & 1453 & 88.7 & 74.9 & 77.2 & 43.0 \\
        \bottomrule
    \end{tabular}
\end{table}

As shown in Table~\ref{tab:mbaq_cross_vlm}, behavior-aligned quantization generalizes across VLM families rather than depending specifically on InternVL3.5.
Nevertheless, the learned quantizer remains backbone-specific, because different VLMs have distinct visual feature spaces and downstream behaviors; a separate MBAQ quantizer should therefore be trained for each target VLM.

%% file: checklist.tex
\section*{NeurIPS Paper Checklist}

\begin{enumerate}

\item {\bf Claims}
    \item[] Question: Do the main claims made in the abstract and introduction accurately reflect the paper's contributions and scope?
    \item[] Answer: \answerYes{} % Replace by \answerYes{}, \answerNo{}, or \answerNA{}.
    \item[] Justification: We did it in the Abstract and Introduction.

\item {\bf Limitations}
    \item[] Question: Does the paper discuss the limitations of the work performed by the authors?
    \item[] Answer: \answerYes{} % Replace by \answerYes{}, \answerNo{}, or \answerNA{}.
    \item[] Justification: We did it in the Conclusion.

\item {\bf Theory assumptions and proofs}
    \item[] Question: For each theoretical result, does the paper provide the full set of assumptions and a complete (and correct) proof?
    \item[] Answer: \answerNA{} % Replace by \answerYes{}, \answerNo{}, or \answerNA{}.
    \item[] Justification: This paper does not include any theoretical result.

    \item {\bf Experimental result reproducibility}
    \item[] Question: Does the paper fully disclose all the information needed to reproduce the main experimental results of the paper to the extent that it affects the main claims and/or conclusions of the paper (regardless of whether the code and data are provided or not)?
    \item[] Answer: \answerYes{} % Replace by \answerYes{}, \answerNo{}, or \answerNA{}.
    \item[] Justification: We disclose all detailed needed to reproduce the main experiments.

\item {\bf Open access to data and code}
    \item[] Question: Does the paper provide open access to the data and code, with sufficient instructions to faithfully reproduce the main experimental results, as described in supplemental material?
    \item[] Answer: \answerYes{} % Replace by \answerYes{}, \answerNo{}, or \answerNA{}.
    \item[] Justification: We will provide open access to the data and code, with sufficient instructions to faithfully reproduce the main experimental results.

\item {\bf Experimental setting/details}
    \item[] Question: Does the paper specify all the training and test details (e.g., data splits, hyperparameters, how they were chosen, type of optimizer) necessary to understand the results?
    \item[] Answer: \answerYes{} % Replace by \answerYes{}, \answerNo{}, or \answerNA{}.
    \item[] Justification: See Experiments and Appendix

\item {\bf Experiment statistical significance}
    \item[] Question: Does the paper report error bars suitably and correctly defined or other appropriate information about the statistical significance of the experiments?
    \item[] Answer: \answerNo{} % Replace by \answerYes{}, \answerNo{}, or \answerNA{}.
    \item[] Justification: Error bars are not reported because it would be too computationally expensive.

\item {\bf Experiments compute resources}
    \item[] Question: For each experiment, does the paper provide sufficient information on the computer resources (type of compute workers, memory, time of execution) needed to reproduce the experiments?
    \item[] Answer: \answerYes{} % Replace by \answerYes{}, \answerNo{}, or \answerNA{}.
    \item[] Justification: We provide it in Experiments and Appendix.
    
\item {\bf Code of ethics}
    \item[] Question: Does the research conducted in the paper conform, in every respect, with the NeurIPS Code of Ethics \url{https://neurips.cc/public/EthicsGuidelines}?
    \item[] Answer: \answerYes{} % Replace by \answerYes{}, \answerNo{}, or \answerNA{}.
    \item[] Justification: Yes, we follow the code of ethics.

\item {\bf Broader impacts}
    \item[] Question: Does the paper discuss both potential positive societal impacts and negative societal impacts of the work performed?
    \item[] Answer: \answerNo{} % Replace by \answerYes{}, \answerNo{}, or \answerNA{}.
    \item[] Justification: This paper focuses on academic research and there’s no explicit negative social impacts.
    
\item {\bf Safeguards}
    \item[] Question: Does the paper describe safeguards that have been put in place for responsible release of data or models that have a high risk for misuse (e.g., pre-trained language models, image generators, or scraped datasets)?
    \item[] Answer: \answerNA{} % Replace by \answerYes{}, \answerNo{}, or \answerNA{}.
    \item[] Justification: This paper poses no such risks.

\item {\bf Licenses for existing assets}
    \item[] Question: Are the creators or original owners of assets (e.g., code, data, models), used in the paper, properly credited and are the license and terms of use explicitly mentioned and properly respected?
    \item[] Answer: \answerYes{} % Replace by \answerYes{}, \answerNo{}, or \answerNA{}.
    \item[] Justification: We strictly followed the licenses.

\item {\bf New assets}
    \item[] Question: Are new assets introduced in the paper well documented and is the documentation provided alongside the assets?
    \item[] Answer: \answerNA{} % Replace by \answerYes{}, \answerNo{}, or \answerNA{}.
    \item[] Justification: This paper does not release new assets.

\item {\bf Crowdsourcing and research with human subjects}
    \item[] Question: For crowdsourcing experiments and research with human subjects, does the paper include the full text of instructions given to participants and screenshots, if applicable, as well as details about compensation (if any)? 
    \item[] Answer: \answerNA{} % Replace by \answerYes{}, \answerNo{}, or \answerNA{}.
    \item[] Justification: This paper does not involve crowdsourcing nor research with human subjects.

\item {\bf Institutional review board (IRB) approvals or equivalent for research with human subjects}
    \item[] Question: Does the paper describe potential risks incurred by study participants, whether such risks were disclosed to the subjects, and whether Institutional Review Board (IRB) approvals (or an equivalent approval/review based on the requirements of your country or institution) were obtained?
    \item[] Answer: \answerNA{} % Replace by \answerYes{}, \answerNo{}, or \answerNA{}.
    \item[] Justification: Our research does not require the IRB.

\item {\bf Declaration of LLM usage}
    \item[] Question: Does the paper describe the usage of LLMs if it is an important, original, or non-standard component of the core methods in this research? Note that if the LLM is used only for writing, editing, or formatting purposes and does \emph{not} impact the core methodology, scientific rigor, or originality of the research, declaration is not required.
    %this research? 
    \item[] Answer: \answerNA{} % Replace by \answerYes{}, \answerNo{}, or \answerNA{}.
    \item[] Justification: We did not use LLMs in the core method development.

\end{enumerate}